\documentclass{article}

 \PassOptionsToPackage{numbers, sort&compress}{natbib}
\usepackage[final]{neurips_2022}

\usepackage[utf8]{inputenc} %
\usepackage[T1]{fontenc}    %
\usepackage{hyperref}       %
\usepackage{url}            %
\usepackage{booktabs}       %
\usepackage{amsfonts}       %
\usepackage{nicefrac}       %
\usepackage{microtype}      %
\usepackage{xcolor}         %

\usepackage{multicol}
\usepackage{multirow}
\usepackage{subfiles}
\usepackage{amssymb}
\usepackage{algpseudocode}
\usepackage[ruled,vlined]{algorithm2e}
\usepackage{amsmath,amsthm,amsfonts,paralist}
\usepackage{tabularx}
\usepackage{thmtools, thm-restate}
\usepackage{graphicx}
\usepackage{textcomp}
\usepackage[shortlabels]{enumitem}
\usepackage[labelformat=simple]{subcaption}

\usepackage{wrapfig}

\newtheorem{definition}{Definition}
\newtheorem{theorem}{Theorem}

\DeclareMathOperator\erf{erf}
\DeclareMathOperator\diag{diag}
\DeclareMathOperator*{\argmax}{arg\,max}
\DeclareMathOperator{\dist}{dist}

\newcommand{\hbs}[1]{\textcolor{black}{#1}}
\newcommand{\rev}[1]{\textcolor{black}{#1}}
\newcommand{\wzm}[1]{\textcolor{black}{#1}}
\newcommand{\wzmb}[1]{\textcolor{black}{#1}}
\newcommand{\wzmc}[1]{\textcolor{black}{#1}}
\newcommand{\wzmd}[1]{\textcolor{black}{#1}}
\usepackage{pifont}%
\usepackage{ upgreek }

\newcommand{\cmark}{\ding{51}}%
\newcommand{\xmark}{\ding{55}}%

\usepackage{todonotes}
\usepackage{threeparttable}

\title{A Coupled Design of Exploiting Record Similarity for Practical Vertical Federated Learning}

\author{
Zhaomin Wu, Qinbin Li, Bingsheng He\\
National University of Singapore\\
\{zhaomin,qinbin,hebs\}@comp.nus.edu.sg
}

\begin{document}

\maketitle

\begin{abstract}

Federated learning is a learning paradigm to enable collaborative learning across different parties without revealing raw data. Notably, \textit{vertical federated learning} (VFL), where parties share the same set of samples but only hold partial features, has a wide range of real-world applications. However, most existing studies in VFL disregard the ``record linkage” process. They design algorithms either assuming the data from different parties can be exactly linked or simply linking each record with its most similar neighboring record. These approaches may fail to capture the key features from other less similar records. Moreover, such improper linkage cannot be corrected by training since existing approaches provide no feedback on linkage during training. In this paper, we design a novel coupled training paradigm, FedSim, that integrates one-to-many linkage into the training process. Besides enabling VFL in many real-world applications with fuzzy identifiers, FedSim also achieves better performance in traditional VFL tasks. Moreover, we theoretically analyze the additional privacy risk incurred by sharing similarities. Our experiments on eight datasets with various similarity metrics show that FedSim outperforms other state-of-the-art baselines. The codes of FedSim are available at \url{https://github.com/Xtra-Computing/FedSim}.
\end{abstract}

\section{Introduction}\label{sec:intro}

\textit{Federated learning} is a collaborative learning framework to train a model from distributed datasets with privacy guarantees. A commonly existing and widely studied scenario of federated learning is \textit{vertical federated learning}~\cite{yang2019federated,li2019survey} (VFL), where multiple parties sharing the same set of samples have different sets of features. We focus on the setting where only one party holds the labels like most of the studies \cite{wu2020privacy,liu2020secure,nock2021impact}. The party holding labels is named \textit{primary party}; the parties without labels are named \textit{secondary parties}. In VFL, the features that exist on multiple parties are called \textit{common features}. The vector of common features in a data record is called the \textit{identifier} of the record.

\wzm{Existing studies \cite{hardy2017private,nock2018entity,nock2021impact,kang2020fedmvt} formulate VFL as two separated processes: linkage and training. In the linkage process, the datasets on different parties are linked according to the identifiers. In the training process, these distributed but linked data records are trained by VFL algorithms. Specifically, in the linkage process of existing studies, each data record is linked to a data record with the exactly matched or the most similar identifier (i.e., \textit{one-to-one} linkage).} \wzmc{Nonetheless, according to our study on \wzmd{all} the completed projects in German Record Linkage Center~\cite{AntoniSchnell2019summary} (GRLC), only 27.3\% of the record-linkage applications are linked on exact identifiers.
\wzmd{The other projects rely on} fuzzy identifiers such as addresses, in which one-to-one linkage can seriously impair the accuracy of the VFL model for the following two reasons.}

\begin{figure*}[t!]
\centering
\begin{subfigure}{0.36\linewidth}
\includegraphics[width=.98\textwidth]{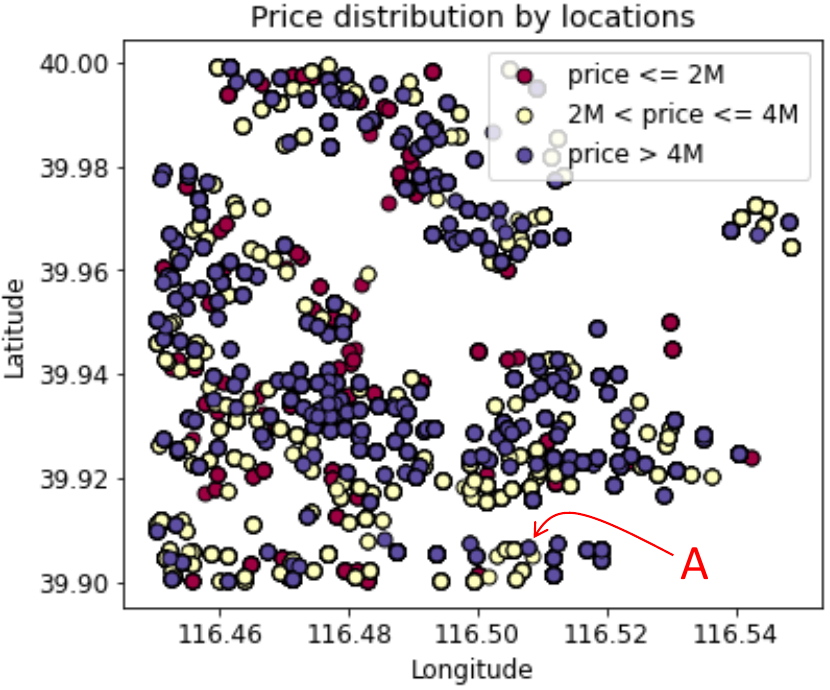}
 \caption{Housing prices by geolocations in Beijing}\label{fig:price_dist}
\end{subfigure}
\hspace{5pt}
\begin{subfigure}{0.6\linewidth}
\includegraphics[width=.98\textwidth]{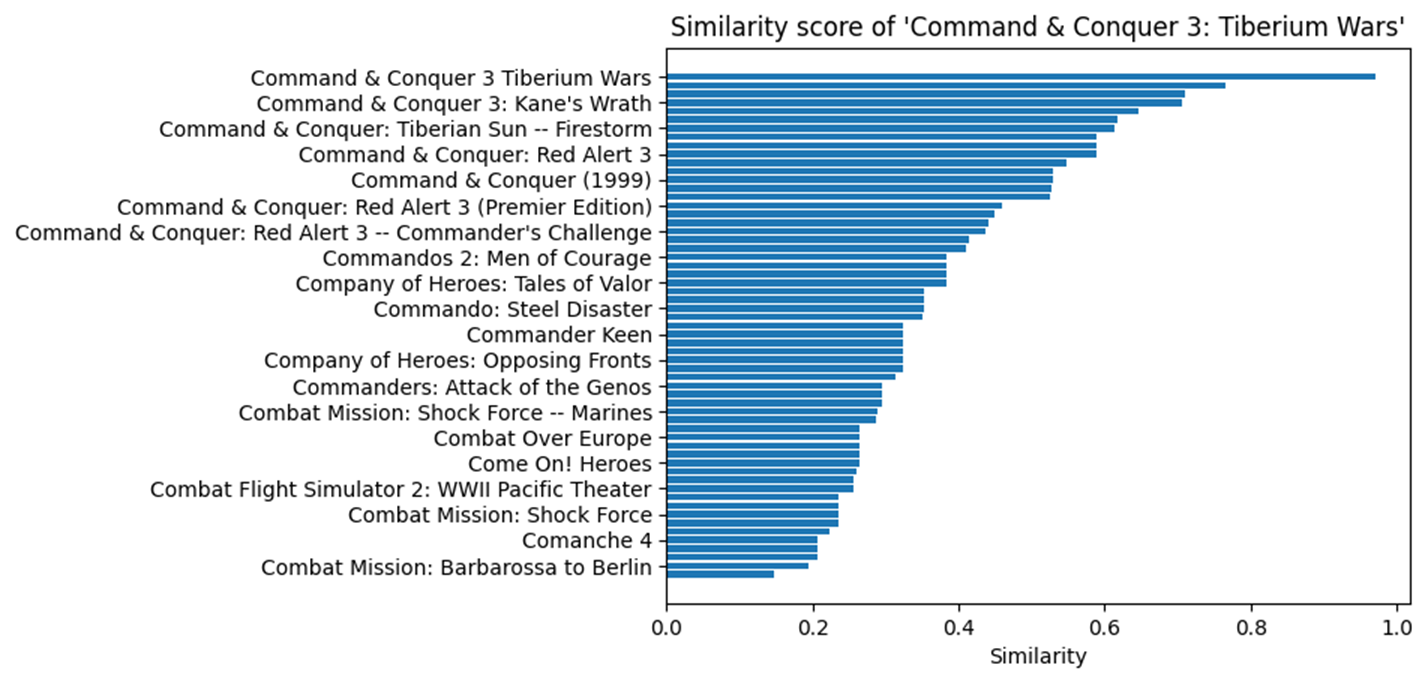}
\caption{Games with similar titles to ``Command \& Conquer 3: Tiberium Wars''}\label{fig:game_dist}
\end{subfigure}
\caption{Examples of real-world record linkage (\texttt{house} and \texttt{game} dataset in the experiments)}\label{fig:example}
\end{figure*}

\wzmd{First, only linking the records with top similarity (i.e., \textit{one-to-one} linkage) does not necessarily capture the key features, which can be demonstrated by two real-world applications. 1) Considering the VFL for price prediction between a real estate company and a house leasing company linked by GPS locations of houses, as shown in Figure~\ref{fig:price_dist}, the closest house to house $A$ (purple) may not be in the same price level as $A$. 2) Considering the VFL between steam games and IGN games linked by the game titles, as shown in Figure~\ref{fig:game_dist}, linking not only the exactly matched game but also other games in the same series intuitively benefits game recommendation tasks. Even in applications where identifiers can be exactly matched (e.g., identifiers are ID), if some features are missing or biased, one-to-one linkage prevents the training process from enhancing these features from other similar records. These applications commonly exist in practice according to our investigation. An opposite extreme case is to link each record with all the records in another party (\textit{one-to-all} linkage), which keeps all the information but is too expensive for both linkage and training. Therefore, a \textit{one-to-many} linkage approach is needed as a balance between efficiency and performance.}

Second, separating the linkage from the training also harms the performance of VFL. In existing VFL approaches, since the linkage process cannot obtain any feedback from the training process, the linkage is conducted with the goal of finding \emph{true-matched} pairs of records instead of finding the pairs that reduce the training loss. Hence, an integrated VFL framework that conducts one-to-many linkage under the guide of training is desired.

To address these two drawbacks, we link each data record to the records with top-$K$ similar identifiers in another party and \wzm{design a coupled framework of linkage and training}. Our main challenge is to effectively exploit these linked pairs and their similarities to boost the performance of VFL. To tackle this challenge, we propose a similarity-based \wzmc{coupled} VFL framework \texttt{FedSim} on the top of SplitNN \cite{vepakomma2018split} which is a VFL algorithm for neural networks. In FedSim, \textit{similarity} is a dominant feature that determines the order and the weight of each pair of linked records. The weights (i.e., impact) of similarities are also adjusted in each training iteration. After the training, each similarity is mapped to a weight based on its contribution to reducing the loss.

Furthermore, to address the additional data privacy issue incurred by FedSim, \wzmc{we first add Gaussian noise to the similarities and analyze the privacy of FedSim under differential privacy~\cite{dwork2014algorithmic}. Our analysis suggests that differential privacy that provides rigorous guarantees for every individual record regardless of the type of attack is impractical for FedSim. As an alternative, we propose an intuitive greedy attack to infer identifiers from similarities, followed by a theoretical bound of the probability of its success.}

Our main contributions can be summarized below. 1) We propose a novel asymmetric training paradigm, named FedSim, and a training-free metric to estimate the improvement of FedSim on baseline approaches; \wzmc{2) we analyze the privacy of FedSim under differential privacy;} 3) we propose a greedy attack on FedSim and the corresponding defense method that can theoretically bound the success rate of this attack; 4) we conduct extensive experiments on three synthetic datasets and five real-world datasets, which indicates that FedSim outperforms state-of-the-art baselines.

\section{Preliminaries}\label{sec:background}
\textbf{Privacy-Preserving Record Linkage.} \textit{Privacy-preserving record linkage} (PPRL) \cite{vatsalan2017privacy} aims to link the data records from two parties that refer to the same sample without revealing real identifiers. Most PPRL methods \cite{vatsalan2017privacy,vatsalan2014scalable,karakasidis2015scalable} consist of three main steps: blocking, comparison, and classification. First, in the blocking step, data records that are unlikely to be linked are pruned to reduce the number of comparisons. Then, in the comparison step, a similarity between identifiers is computed for each candidate pair of data records. Finally, in classification, each candidate pair is classified as ``matches'' or ``non-matches'', which is usually done by a manually set threshold. Notably, although blocking ensures sample scalability, party scalability remains a challenge in PPRL. Therefore, we focus on the linkage and training process of two parties and provide an extension to multiple parties in this paper.

Although the training of FedSim does not rely on any specific PPRL framework, our privacy analysis is based on a state-of-the-art PPRL framework FEDERAL~\cite{karapiperis2017federal}. Coordinated by an honest-but-curious server, FEDERAL calculates similarities by comparing Bloom filters \cite{vatsalan2017privacy} generated from identifiers. It is theoretically guaranteed that all the identifiers generate Bloom filters containing similar numbers of ones; thus, attackers are hard to distinguish identifiers based on Bloom filters. \wzmc{More detailed background of FEDERAL is introduced in Appendix~\ref{apdx:background}.}

\textbf{Vertical Federated Learning.}
In this paragraph, we present a formal definition of VFL between two parties. Suppose two parties $P$ and $S$ want to cooperate with each other to train a machine learning model. $P$ is the primary party holding $m$ samples and labels $\{\mathbf{x}^P,\mathbf{y}\}\triangleq\{x^P_i,y_i\}_{i=1}^m$, $S$ is the secondary party holding $n$ samples $\mathbf{x}^S\triangleq\{x^S_i\}_{i=1}^n$ which can also benefit the machine learning task. In order to perform linkage, we assume there are some common features between $\{x^P_i\}_{i=1}^m$ and $\{x^S_i\}_{i=1}^n$, i.e., $\{x^P_i\}_{i=1}^m=\{d^P_i,k^P_i\}_{i=1}^m, \{x^S_i\}_{i=1}^n=\{d^S_i,k^S_i\}_{i=1}^n$, where $k^P_i,k^S_i$ are common features used for linkage and $d^P_i,d^S_i$ are remaining features used for training with dimension $l_P,l_S$. We denote $\mathbf{d}^P\triangleq\{d^P_i\}_{i=1}^m$, $\mathbf{d}^S\triangleq\{d^S_i\}_{i=1}^n$ for simplicity. %
Our goal is to enable party P to exploit $\mathbf{x}^S$ to train a model that minimizes the global loss. Formally, we aim to optimize the following formula:
\[
\min_\theta \frac{1}{m}\sum_{i=1}^{m}L(f(\theta;x^P_i,\mathbf{x}^S);\;y_i)+\lambda\Omega(\theta)
\]where $L(\cdot)$ is the loss function, $f(\cdot)$ is the VFL model, and $\lambda\Omega(\theta)$ is the regularization term.

\begin{figure}
\centering
\begin{subfigure}{0.53\linewidth}
\includegraphics[width=.99\linewidth]{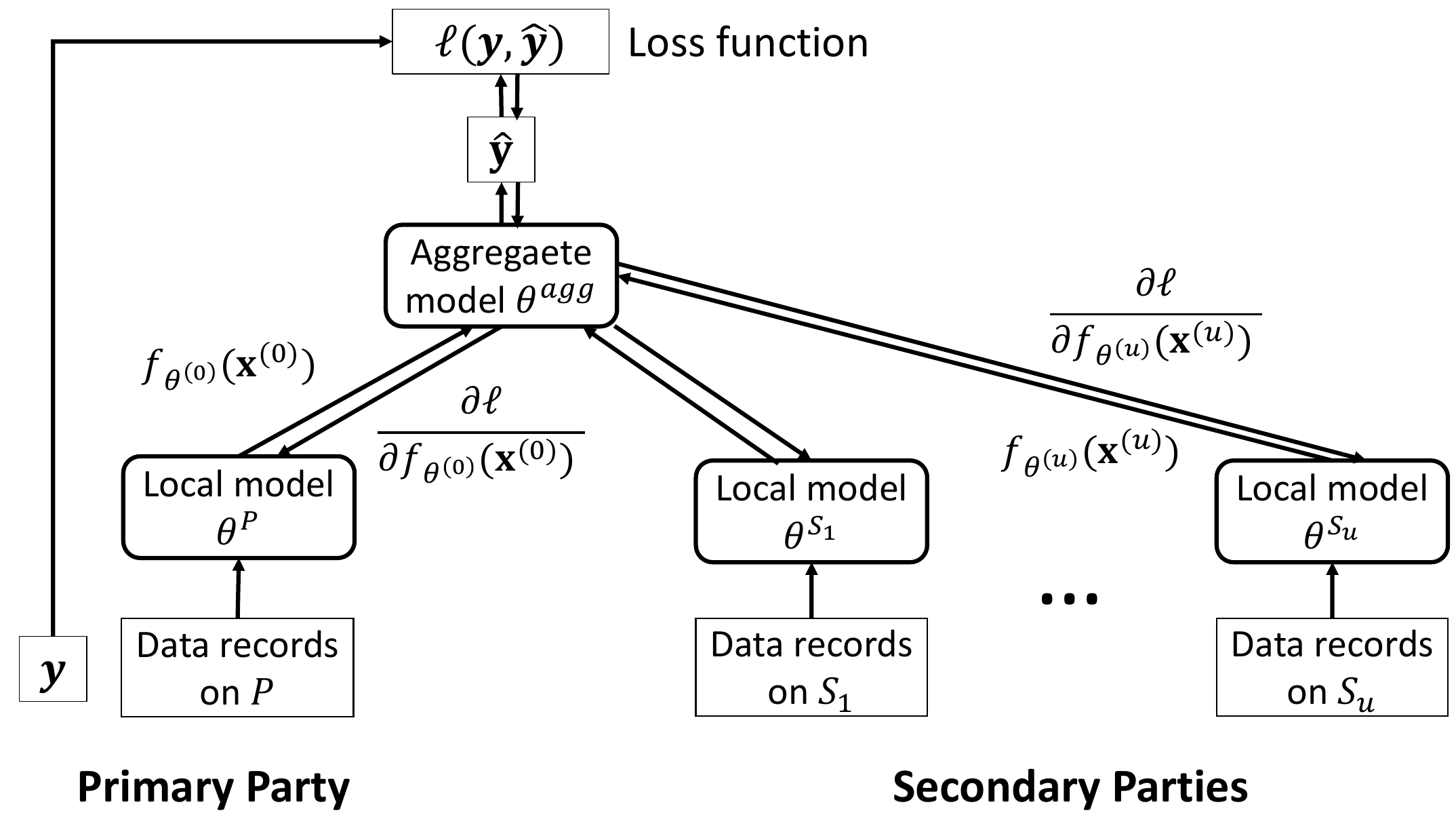}
\caption{\rev{Forward and back-propagation of SplitNN}}\label{fig:splitnn}
\end{subfigure}
\begin{subfigure}{0.46\linewidth}
\includegraphics[width=.99\linewidth]{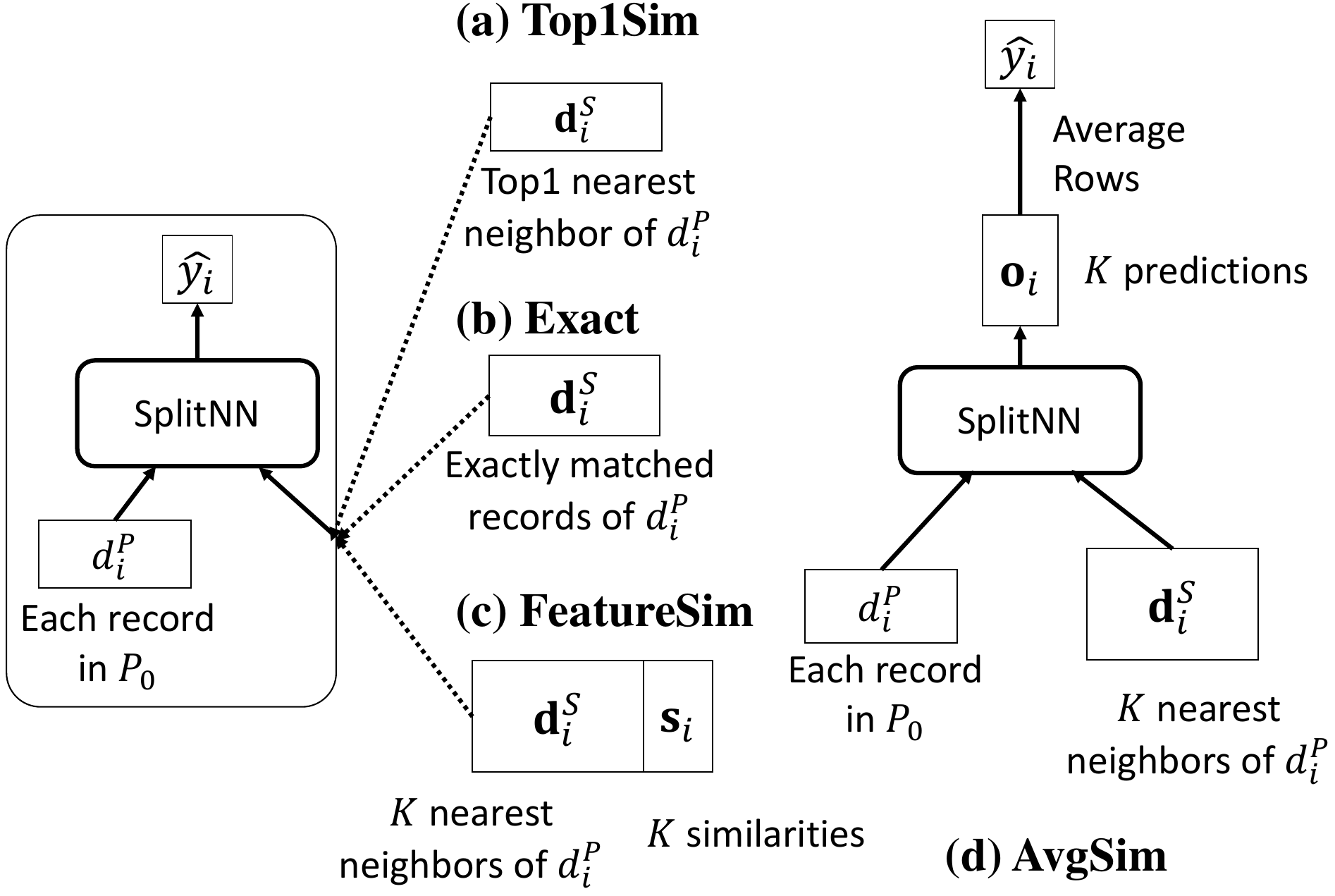}
    \caption{Structure of baseline models} \label{fig:baseline}
\end{subfigure}
\caption{Structure of SplitNN and baseline VFL models}
\end{figure}

\textbf{SplitNN.} Most vertical federated learning algorithms only support simple models like logistic regression \cite{hu2019learning} which are ineligible to handle many real-world applications. Therefore, we adopt SplitNN~\cite{vepakomma2018split} which is a popular and state-of-the-art VFL algorithm that supports neural networks. The main idea of SplitNN is to split a model into multiple parties and conduct training by transferring gradients and intermediate outputs across parties. As shown in Figure~\ref{fig:splitnn}, a global model is split into an aggregation model $\theta^{agg}$ and $H$ local models $\theta^P,\{\theta^{S_u}|u\in[1,H-1]\}$, where $H$ is the number of parties. In each iteration, the outputs of local models are calculated by forward propagation and sent to the primary party $P_0$ which holds the labels. After concatenating the outputs of local models, $P$ continues forward propagation to derive the final prediction $\hat{y}$ and the loss $\ell$. Then, $P$ performs back-propagation until the input of the aggregation model. The gradients w.r.t. the outputs of each local model are calculated and sent to the corresponding secondary party. Finally, all $H$ parties finish back-propagation with the gradients. %

\section{Baseline Approaches}\label{subsec:baseline}

By extensively reviewing the existing VFL approaches, we find that all the existing approaches only use exactly matched (Exact) or most similar (Top1Sim) pairs of data records in the training. We denote these approaches that separate the linkage and training and as \textit{separated approaches}. \wzmc{According to our \wzmd{detailed} study in Appendix~\ref{apdx:application}, only 27.3\% of the record-linkage applications in GRLC based on exact identifiers are suitable for these approaches.} To design a \wzm{\textit{coupled approach}} that effectively exploits linkage information, we analyze the main drawback of existing approaches and some baseline coupled frameworks.

\textbf{Exact~\cite{cheng2019secureboost,liu2020federated,liu2020asymmetrically,wu2020privacy,fu2021vf2boost,wenjie2021vertical,hu2019fdml}/Top1Sim~\cite{nock2018entity,kang2020fedmvt,hardy2017private, nock2021impact}.} (Figure~\ref{fig:baseline}(a,b)) These baseline \wzmc{separated} approaches that only link exactly matched or most similar pairs neglect information of less similar but useful pairs, resulting in a poor performance of VFL model \wzmd{in many real applications such as those in GRLC.}

\textbf{AvgSim.} (Figure~\ref{fig:baseline}(d)) We propose AvgSim as a baseline \wzmc{coupled} approach that considers multiple pairs of records. Specifically, each data record in $x^P_i$ party P is linked with its $K$ most similar data records in party S. The prediction of $x^P_i$ is an average of prediction when linking $x^P_i$ with its $K$ nearest $x^S_j$. Though considering more pairs, AvgSim overemphasizes the pairs with medium similarities, leading to redundant noise added to the model. Unaware of the similarity of each pair, the model is unable to filter out this redundant noise.

\textbf{FeatureSim.} (Figure~\ref{fig:baseline}(c)) We propose FeatureSim as a \wzmc{coupled} baseline approach that adopts similarities to the training. Each $x^P_i$ in party P is linked with its $K$ most similar data records in party S, and the similarity between each pair is appended to the record. Nonetheless, similarity, which contains critical linkage information, is treated equally to other features. This drawback limits the ability of VFL models to extract information from similarities, thus affecting the overall performance. Such an impact can be significant according to our experiments.

\wzmc{Based on the above analyses, we clearly observe that the record linkage should be coupled with the design of VFL by taking advantage of record similarity not only during the linkage procedure but also during the training procedure. Meanwhile, according to the analysis on AvgSim, more advanced models are needed to merge the outputs of linked pairs with $K$ largest similarities. These analyses motivate our design of FedSim as a coupled framework of linkage and training.} 

\section{Our Approach: FedSim}\label{sec:approach}
Our approach has two components: soft linkage and similarity-based VFL. In soft linkage, after finding top-$K$ similar pairs by existing PPRL methods, the server preprocesses the similarities by normalization and Gaussian noise perturbation. The scale of noise can be determined by a constant bound $\tau$ of the attacker's success rate, which is further analyzed in Section~\ref{sec:privacy}. Then, the aligning information and aligned similarities are sent to party P or S which will align the data records accordingly. In similarity-based VFL, we design a model (as shown in Figure~\ref{fig:training}) by adding additional components around SplitNN to effectively exploit similarities. Finally, this model, taking the aligned data records and aligned similarities as input, is trained by back-propagation like SplitNN. \hbs{Although we use SplitNN as an example, the idea of soft linkage and similarity-based learning can be extended to other federated learning algorithms.}

\subsection{Soft Linkage}\label{subsec:pprl}

\wzmb{Soft linkage, following traditional PPRL in ``blocking'' and ``comparison'' steps, directly outputs the similarities with normalization and noise addition without performing the ``classification'' step.} Same as many existing PPRL approaches~\cite{karakasidis2015scalable,karapiperis2017federal,vatsalan2016privacy,karapiperis2014lsh}, we assume there exists an honest-but-curious server coordinating the linkage process. \wzmb{Specifically,} soft linkage, taking $\mathbf{k}^P$ and $\mathbf{k}^S$ as input, outputs the alignment information (i.e., indices that indicate the order of records) and similarities. First, each sample $x_i^P$ in party P is linked with samples containing $K$ most similar identifiers in party S. The similarities between identifiers of these linked pairs are calculated by a PPRL protocol. To fully utilize $\mathbf{x}^S$ for each sample $x^P_i$, $K$ should be large enough to ensure that all the pairs which may benefit the model are included. \wzm{Notably, setting a large $K$ hardly leads to overfitting according to our experiments in Appendix~\ref{apdx:choose_k}}. Second, the server calculates the raw similarities $\rho_{ij}$ between these linked pairs; the raw similarity $\rho_{ij}$ is defined as normalized negative distance. Formally,
\begin{equation}\label{eq:calc_sim}
    \rho_{ij}=\frac{-\dist(k^P_i,k^S_j)-\mu_0}{\sigma_0}.
\end{equation}
where $\mu_0$ and $\sigma_0$ are the mean and standard variance of all negative distances $-\dist(k^P_i,k^S_j)$. To prevent the attacker from guessing the vectors $k^P_i$ or $k^S_j$ from similarities (further discussed in Section~\ref{sec:privacy}), we add Gaussian noise of scale $\sigma$ to each $\rho_{ij}$. Formally, for $\forall i\in[1,m], \forall j\in[1,K]$
\begin{equation}\label{eq:add_noise}
    s_{ij} = \rho_{ij}+N(0,\sigma^2).
\end{equation}
For simplicity, we denote $\mathbf{s}_i\triangleq\{s_{ij}\}_{j=0}^K$ and $\mathbf{s}\triangleq\{\mathbf{s}_{i}\}_{i=0}^m$.

After the similarities are calculated, the server directly sends the similarities $\mathbf{s}$ to party P and sends the aligning information (i.e., indices that indicate orders) to both parties. Finally, parties P and S align their samples or similarities according to the aligning information to ensure that each $x^P_i$, $\mathbf{x}^S_i$ and $\mathbf{s}_i$ refer to the same sample.

\subsection{Similarity-Based VFL}\label{subsec:sim_vfl}

\begin{wrapfigure}{r}{0.5\textwidth}
    \includegraphics[width=\linewidth]{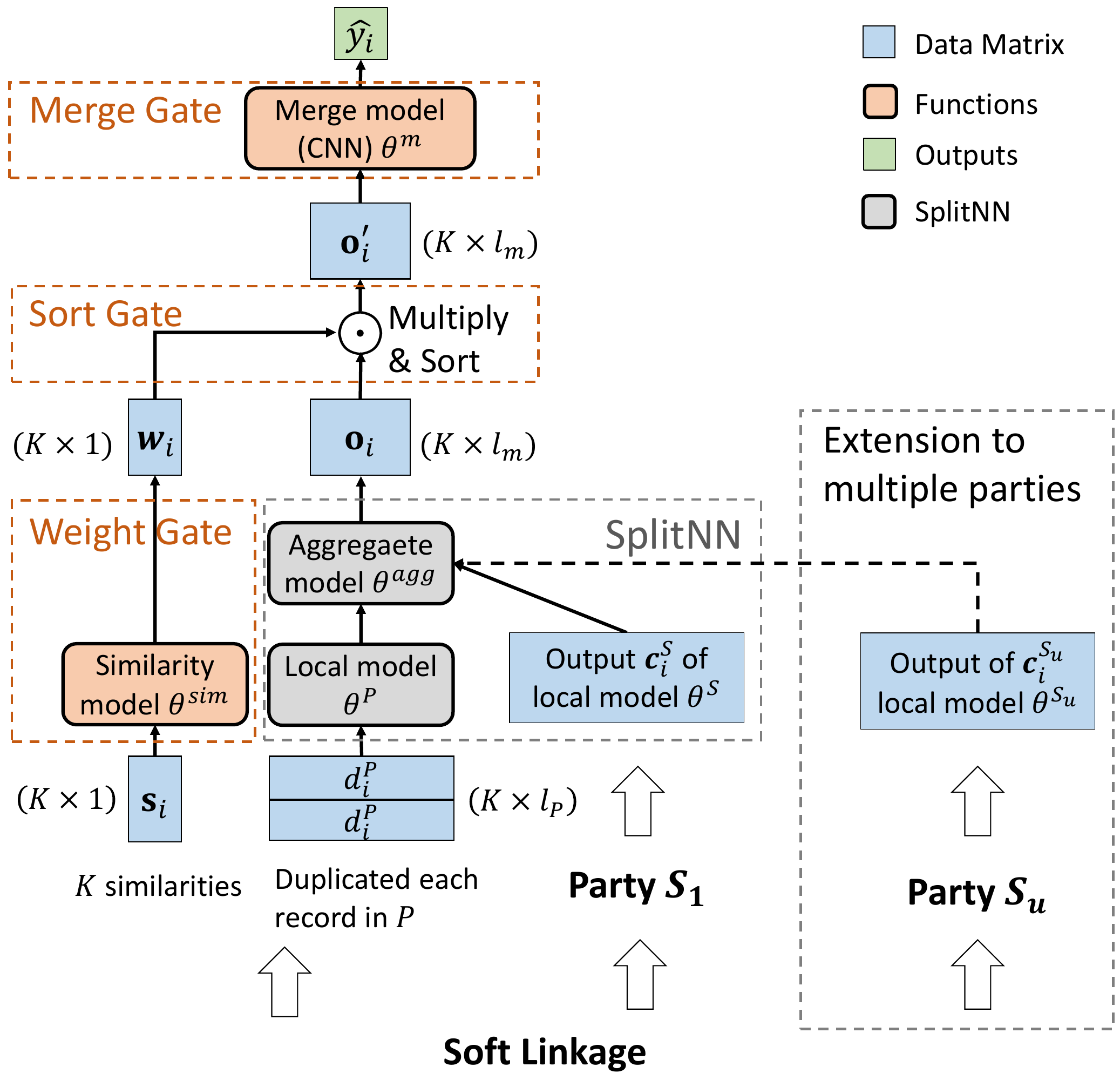}
    \caption{Model structure of FedSim}\label{fig:training}
\end{wrapfigure}

The training process is summarized in Algorithm~\ref{alg:main} and the model structure is shown in Figure~\ref{fig:training}. As discussed in Section~\ref{sec:related_work}, FedSim is designed on top of SplitNN~\cite{vepakomma2018split}, which makes preliminary predictions $\mathbf{o}_i$ (with $l_m$ dimensions) for each $d_i^P$ and its $K$ neighbors $\mathbf{d}^S_i$ (lines 4-7). Specifically, we 
add three \textit{gates} (weight gate, merge gate, sort gate) around SplitNN and train the whole model similarly to SplitNN. The main function of these gates is to effectively exploit similarities $\mathbf{s}_i$ to merge the preliminary outputs $\mathbf{o}_i$ into the final prediction $\hat{y}_i$.

\textbf{Weight Gate.}  One straightforward idea is that more similar pairs of samples contribute more to the performance. Thus, the rows in $\mathbf{o}_i$ with higher similarities should be granted a larger weight. Directly multiplying the similarities to $\mathbf{o}_i$ is inappropriate since the values of similarities do not directly represent the importance of records. Therefore, in weight gate, we use \textit{similarity model}, a simple neural network with one-dimensional input and one-dimensional output, to \wzm{non-linearly} map the similarities to weights $\mathbf{w}_i$ which indicates the importance of the record (line 8). Then, the weighted outputs are calculated as a matrix multiplication (line 9).

\textbf{Merge Gate.}  The merge gate contains a merge model to aggregate the information in $\mathbf{o}_i'$ (line 11). For tasks that only require a linear aggregation, setting the merge model as an ``average over rows'' of $\mathbf{o}_i'$ is sufficient to obtain a promising result. However, for tasks that require a non-linear aggregation, a neural network is required to be deployed as a merge model. In this case, although the merge model can be implemented by a multi-layer perceptron (MLP) which takes flattened $\mathbf{o}_i'$ as input, such an approach usually leads to over-fitting due to a large number of parameters. Meanwhile, the flatten operation causes information loss because the merge model is unaware of which $K$ features in the flattened $\mathbf{o}_i'$ correspond to the same output feature in $\mathbf{o}_i'$. Considering these two factors, we set the merge model as a 2D convolutional neural network (CNN) with kernel size $k_{conv}\times1$. \wzmb{$\mathbf{o}_i'$, containing $K$ output vectors with $l_m$ dimensions, is regarded as a 2D input of size $K\times l_m$ for the CNN merge gate. Then, the merge gate effectively merges these output vectors with close similarities. }

\textbf{Sort Gate.}  The sort gate is an optional but sometimes crucial module depending on the property of the chosen merge model. For merge models that are insensitive to the order of $\mathbf{o}_i'$ (e.g., averaging over rows), sorting is not needed because the order does not affect the output of the merge model. However, for merge models that are sensitive to the order of $\mathbf{o}_i'$ (e.g., neural networks), inconsistent order of features can incur irregular sharp gradients which makes the merge model hard to converge. Thus, $\mathbf{o}_i'$ should be sorted by similarities (line 10) to stabilize the updates on the merge model. Also, grouping the pairs with close similarities together helps the merge gate effectively aggregate the information.

\begin{algorithm}[h!]
\LinesNumbered
\SetKwInOut{Input}{Input}
\SetKwInOut{Output}{Output}
\small
\Input{Aligned datasets and labels $\mathbf{d}^P$, $\mathbf{d}^S$, $\mathbf{y}$; similarities $\mathbf{s}$; number of similar samples $K$; number of epochs $T$; number of samples $m$ in party P; }
\Output{SplitNN parameter $\theta^P_t,\theta^S_t,\theta^{agg}$; similarity model parameter $\theta^{sim}_t$; merge model parameter $\theta^m_t$}
Initialize $\theta^P_0,\theta^S_0,\theta^{sim}_0,\theta^m_0,\theta^{agg}_0$;\tcp*[f]{FP: forward propagation; BP: back-propagation}\\
\For{$t\gets 0$ \KwTo $T$}{
    \For{$i\gets 0$ \KwTo $m$}{
        \textbf{Party S} loads $i$-th batch $\mathbf{d}^S_i$ from $\mathbf{d}^S$\;
         calculates $\mathbf{c}^S_i=f(\theta^S_t;\mathbf{d}^S_i)$ and sends $\mathbf{c}^S_i$ to party P; \tcp*[f]{Local model FP}\\
        
        \textbf{Party P} loads $i$-th sample $d^P_i$ from $\mathbf{d}^P$\;
         receives $\mathbf{c}^S_i$ from party S and calculates $\mathbf{o}_i=f(\theta^P_t;\mathbf{c}^S_i,d^P_i)$;\tcp*[f]{SplitNN FP}\\
         loads similarities $\mathbf{s}_i$ from $\mathbf{s}$ and calculates $\mathbf{w}_i=f(\theta^{sim}_t;\mathbf{s}_i)$;\tcp*[f]{Similarity model FP}\\
         calculates weighted outputs $\mathbf{o}_i'=\diag(\mathbf{w}_i)\mathbf{o}_i$\;
         sorts the rows of $\mathbf{o}_i'$ by $\mathbf{s}_i$ (or $\mathbf{w}_i$);\tcp*[f]{Sort gate}\\
         calculates $\hat{y_i}=f(\theta^m_t;\mathbf{o}_i')$ with sorted $\mathbf{o}_i'$;\tcp*[f]{Merge model FP}\\
         calculates gradients $\mathbf{g}^P_t=\nabla_{\theta^P_t} L(\hat{y_i},y_i)$, $\mathbf{g}^{sim}_t=\nabla_{\theta^{sim}_t} L(\hat{y_i},y_i)$,  $\mathbf{g}^m_t=\nabla_{\theta^m_t} L(\hat{y_i},y_i)$, $\mathbf{g}^c_t=\nabla_{\mathbf{c}^S_i} L(\hat{y_i},y_i)$ and sends $\mathbf{g}^c_t$ to party S;\tcp*[f]{Merge and similarity model BP}\\
         updates parameters $\theta^{agg}_{t+1}=\theta^{agg}_{t}-\eta_t\theta^{agg}_{t},\theta^P_{t+1}=\theta^P_{t}-\eta_t \mathbf{g}^P_{t}, \theta^{sim}_{t+1}=\theta^{sim}_{t}-\eta_t \mathbf{g}^{sim}_{t}, \theta^m_{t+1}=\theta^m_{t}-\eta_t \mathbf{g}^m_{t}$\;
        
        \textbf{Party S} receives $\mathbf{g}^c_t$ from party P and continues calculating gradients $\mathbf{g}^S_t=\nabla_{\theta^S_t}\mathbf{g}^c_t$\;
        updates parameters $\theta^S_{t+1}=\theta^S_{t}-\eta_t \mathbf{g}^S_{t}$;\tcp*[f]{Local model BP}\\
    }
}
\caption{Training Process of FedSim}\label{alg:main}
\end{algorithm}

The whole model with SplitNN and three gates performs back-propagation by transferring gradients like SplitNN (lines 12, 14). After gradients are calculated, all the parameters are updated by gradient descent (lines 13, 15). \wzm{According to the experiment in Appendix~\ref{subsec:train_time}, FedSim incurs longer but acceptable training time compared to Exact and Top1Sim.} \wzmb{During the inference, for each data record in the primary party, $K$ most similar data records in secondary parties are linked. These linked data records are fed into the FedSim model to derive the final prediction.} \wzmd{More insights into the weight gate and merge gate are elaborated by visualization in Appendix~\ref{subsec:visualization}.}

\subsection{Improvement Estimation}
\begin{wrapfigure}{r}{0.32\textwidth}
    \includegraphics[width=\linewidth]{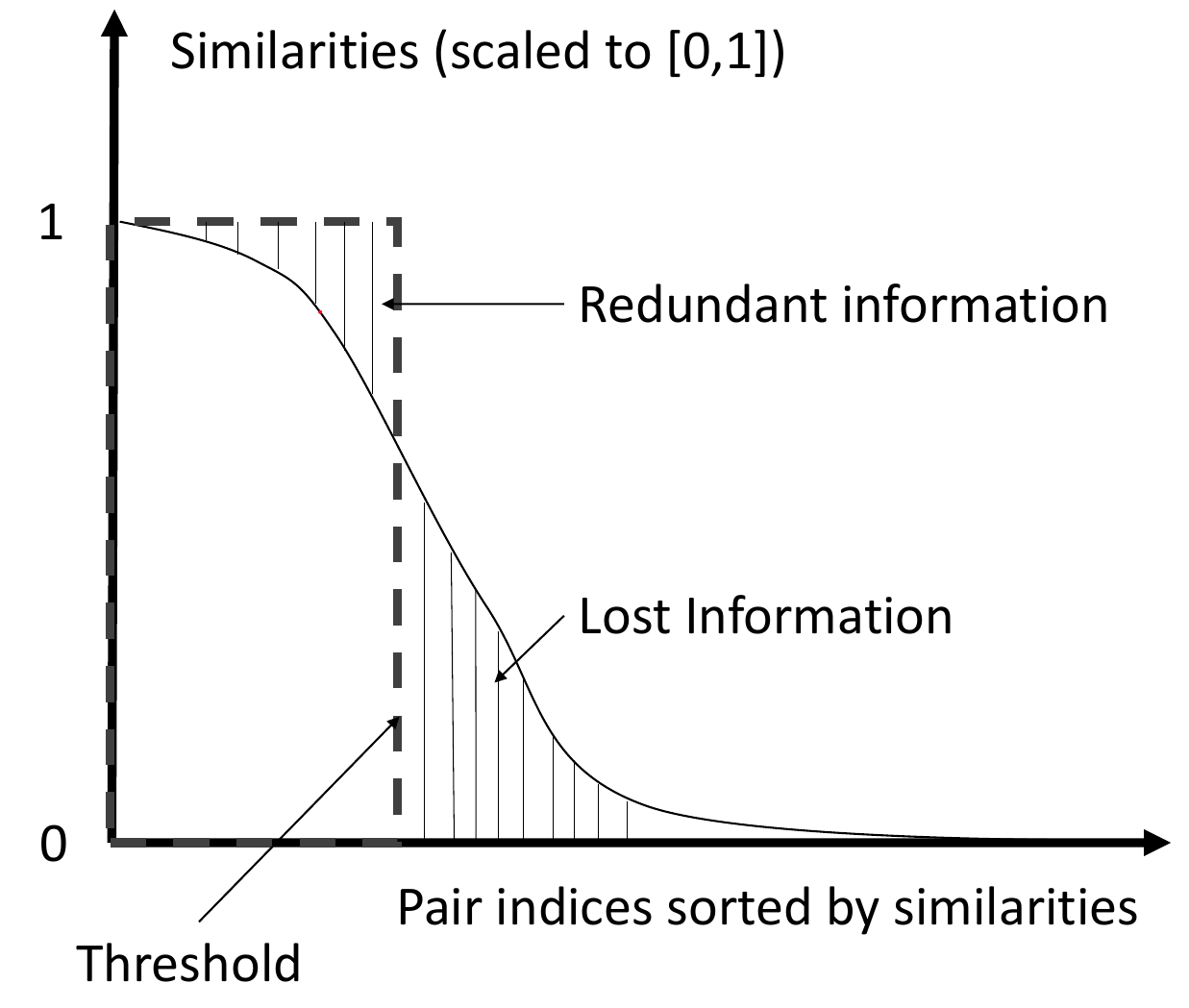}
    \caption{Estimated improvement of FedSim on AvgSim with top-K neighbors}
    \label{fig:metric}
\end{wrapfigure}

Similar to many existing studies ~\cite{cheng2019secureboost,hardy2017private,nock2021impact,wu2020privacy} in VFL, we mainly focus on the two-party setting which has many real-world applications (e.g., bank and fintech company \cite{wu2020privacy}). Meanwhile, we also support an extension to the multiple-party setting as elaborated in Appendix~\ref{sec:multi_party}.

In this subsection, we propose a data-linkage-based metric to estimate the improvement of FedSim over baselines (i.e., AvgSim, Top1Sim, Exact) without training. Calculating all the similarities between $x^P_i$ and every $\{x^S_j\}_{j=1}^n$, we can plot the sorted similarities of each pair as a curve. For example, Figure~\ref{fig:metric} displays the curve of AvgSim with top-K neighboring records. The performance of these baselines is impeded by two main factors: 1) \textit{lost information} - some pairs of samples with small similarities are neglected. For example, in Figure~\ref{fig:metric}, the samples above the threshold are neglected; 2) \textit{redundant information} - some pairs of samples with smaller similarities are treated equally to those pairs with large similarities. For example, in Figure~\ref{fig:metric}, some samples below the threshold with medium similarities are overestimated. To jointly estimate these two factors, we intuitively assume the information contained in each pair is proportional to its similarity; then, the lost information and the redundant information can be estimated by the similarities. We formally define the metric as follows.

\begin{definition}
Given a data linkage between $\mathbf{x}^P\triangleq\{x^P_i\}_{i=1}^m$ and $\mathbf{x}^S\triangleq\{x^S_j\}_{j=1}^n$, for a data record $x^P_i$ in P, a VFL algorithm $\mathcal{F}$ divide the indices of $n$ records in S into matches and non-matches, denoted as $S_{\text{matches}}$ and $S_{\text{non-matches}}$, respectively. Denote $s_{ij}$ as the scaled similarity between $x^P_i$ and $x^S_j$, the improvement of FedSim on $\mathcal{F}$ for $x^P_i$ is defined as $\Delta_i(\mathcal{F})\triangleq\sum_{j\in S_\text{matches}}(1-s_{ij})+\sum_{j\in S_{\text{non-matches}}s_{ij}}$; the overall improvement of FedSim on $\mathcal{F}$ is defined as $\Delta(\mathcal{F})=\frac{1}{m}\sum_{i=1}^{m}\Delta_i(\mathcal{F})$.
\end{definition}

As shown in Figure~\ref{fig:metric_improve}, this metric can effectively estimate the improvement of FedSim compared with the other baselines in the experiments.

\section{Privacy}\label{sec:privacy}

The shared information in FedSim includes:
\begin{enumerate}[(1)]
    \item similarities $\mathbf{s}$  shared to party P;
    \item intermediate results in SplitNN shared to party P;
    \item intermediate results in PPRL shared to the server.
\end{enumerate}
The intermediate results in SplitNN and PPRL are respectively studied in \cite{vepakomma2020nopeek} and \cite{karapiperis2017federal} (see Section~\ref{sec:background} for details), both of which are  orthogonal to this paper. Therefore, we mainly study the privacy risk caused by similarities $\mathbf{s}$. 

\subsection{Differential Privacy}\label{subsec:priv_analyze}

According to Equation~\ref{eq:calc_sim} and \ref{eq:add_noise}, each similarity $s_{ij}$ is calculated from $k^P_i$ and $k^S_j$. \wzmc{We formally define this procedure as $\mathcal{G}$ below.}

\textbf{Procedure $\mathcal{G}$:} \wzmc{Take $\mathbf{k}^S$ as the input. Each $k_i^P\in \mathbf{k}^P$ is linked to multiple $k_j^S\in \mathbf{k}^S$; the similarities between $k_i^P$ and $k_j^S$ are calculated by $s_{ij}=-\frac{\|k_i^P-k_j^S\|_2+\mu_0}{\sigma_0}+N(0,\sigma^2)$, where $N(0,\sigma^2)$ refers to Gaussian distribution with variance $\sigma^2$. Output the similarities $\mathbf{s}$.}

\begin{restatable}{theorem}{dpFedSimAdv}\label{thm:dp_sim_adv}
    \wzmc{Suppose $\varepsilon > 0$ and $0 < \delta <1/2-e^{-3\varepsilon}/\sqrt{2\pi\varepsilon}$. Suppose the size of $\mathbf{k}^P$ is $n\times \beta$. If procedure $\mathcal{G}$ is $(\varepsilon,\delta)$-DP, then $\sigma\ge \Delta_\mathcal{G}/\sqrt{2\varepsilon}$, where $\Delta_\mathcal{G}=n\cdot \max\left\{\left|\frac{1+\mu_0}{\sigma_0}\right|,\left|\frac{-1+\mu_0}{\sigma_0}\right|\right\}$. }
\end{restatable}

\wzmc{The proof of Theorem~\ref{thm:dp_sim_adv} is included in Appendix~\ref{subsec:proof_dp}. As observed from Theorem~\ref{thm:dp_sim_adv}, the noise scale $\sigma$ increases linearly by $n$. The large scale of noise derived from differential privacy would seriously affect the performance of FedSim. Hence, we further analyze the privacy risk of FedSim against specific attacks.}

\subsection{Privacy Against Greedy Attacks}

In this subsection, our analysis focuses on a greedy attacker who first predicts the most likely distance from each $s_{ij}$ and then predicts the most likely Bloom filter from the predicted distance. Assuming the attacker already knows the scaling parameters $\mu_0,\sigma_0$, we formulate the attack method as follows.

\textbf{Attack Method.} To obtain $\hat{k}^S_j$ as a prediction of $k^S_j$, the attacker
1) predicts a set of normalized negative distances $\hat{\rho}_{ij}\;(i\in Q)$ from $s_{ij}\;(i\in Q)$, respectively, by maximum a posteriori (MAP) estimation with Gaussian prior $N(0,1)$, i.e., $\hat{\rho}_{ij}=\argmax_{{\rho}_{ij}} p(\rho_{ij}|s_{ij})$;
2) calculates each distance $\hat{u}_{ij}$ by scaling back $\hat{\rho}_{ij}$ with parameters $\mu_0,\sigma_0$, i.e., $\hat{u}_{ij}=-\sigma_0\hat{\rho}_{ij}-\mu_0\;(i\in Q)$;
3) uniformly guesses $\hat{k}^S_j$ from all possible values satisfying $\forall i\in Q, \dist(k^P_i,\hat{k}^S_j)=\hat{u}_{ij}$ which have the same probability of being the real $k^S_j$.

Besides the greedy attack, advanced attackers may predict through the probability distribution of distances rather than predict through the most likely distance. Some attackers may even know some side information like the prior distribution of $k^S_j$ and employ this side information to launch attacks. These advanced attacks are further discussed in Appendix~\ref{sec:adv_attack}.

\wzmc{Although not satisfying differential privacy, we prove that the success probability of greedy attacks }is always bounded by a small constant related to $\sigma_0$ and $\sigma$ regardless of the choice of $Q$ if an attacker follows the attack method (Theorem~\ref{thm:main}).

\begin{restatable}{theorem}{main}\label{thm:main}
    Given a finite set of perturbed similarities \(s_{ij}\;(i\in Q)\) between \(|Q|\) bloom-filters \(k^P_{i}\;(i\in Q)\) in party P and one Bloom filter \(k^S_j\) in party S, if an attacker knows the scaling parameters \(\mu_0,\sigma_0\) and follows the procedure of the attack method, the probability of the attacker's predicted Bloom filter \(\hat{k}^S_j\) equaling the real Bloom filter \(k^S_j\) is bounded by a constant $\tau$. Formally,
	\[\small
	\Pr\left[\left.\hat{k}^S_j=k^S_j\right|\{s_{ij}|i\in Q\},\{k^P_{i}|i\in Q\},\mu_0,\sigma_0,\mathcal{A}\right]\le\tau
	\]
	where constant $\tau=\erf\left(\frac{\sqrt{\sigma^2+1}}{2\sqrt{2}\sigma\sigma_0}\right)$; \(\erf(\cdot)\) is the error function, i.e.,
	\(
	\erf(x)=\frac{2}{\sqrt{\pi}}\int_{0}^{x}{e^{-t^2}}dt
	\); event $\mathcal{A}$: attackers follow the given attack method.
\end{restatable}

The proof of Theorem~\ref{thm:main} is included in Appendix~\ref{sec:proof}. From Theorem~\ref{thm:main}, we find two factors that affect the attacker's success rate: 1) noise added to the similarity ($\sigma$); 2) standard variance of Bloom filters ($\sigma_0$). Among these factors, $\sigma_0$ determines the lower bound of the success rate because $\sqrt{\sigma^2+1}/(2\sqrt{2}\sigma\sigma_0) < 1/(2\sqrt{2}\sigma_0)$, and $\sigma$ determines how close success rates FedSim can guarantee compared to the lower bound. When $\sigma$ is large enough, increasing $\sigma$ helps little with reducing the success rate. Therefore, to ensure the good privacy of FedSim, we should first guarantee a large enough variance among the Bloom filters and then add a moderate noise to the similarities.
\wzmc{
Taking \textit{house} dataset (see Section~\ref{subsec:exp_setup}) as an example where $\mu_0=-46237.78, \sigma_0=21178.86$. Letting $\delta=10^{-5}, \sigma=0.4$, the differential privacy parameter $\varepsilon=2.96\times 10^9$ implies that there is almost no differential privacy guarantee at all. Nonetheless, the attacking success rate $\tau=1.94\times 10^{-5}$ suggests that the privacy risk from certain attacks is very low. Considering the 19479 samples in the training set of party S, only 0.378 (\wzmd{smaller} than one) Bloom filters are expected to be disclosed.}

\section{Experiment}\label{sec:experiment}
\subsection{Experimental Setup}\label{subsec:exp_setup}

\textbf{Dataset.}
We evaluate FedSim on three synthetic datasets (\texttt{sklearn} \cite{sklearn}, \texttt{frog} \cite{frog}, \texttt{boone}, \cite{boone}) and five real-world datasets (\texttt{house}~\cite{house,airbnb}, \texttt{taxi} \cite{taxi,bike}, \texttt{hdb} \cite{hdb,school}, \texttt{game} \cite{steam,rawg}, and \texttt{company} \cite{bertin2011million,defferrard2016fma}). \wzm{The details (e.g. dimension of identifiers) of these datasets are summarized in Appendix~\ref{sec:exp_detail}}. For each real-world dataset, we collect two public datasets from different real-world parties and conduct VFL on both public datasets. For each synthetic dataset, we first create a global dataset by generating with sklearn API \cite{sklearn} (\texttt{sklearn}) or collecting from public (\texttt{frog}, \texttt{boone}). Then, we randomly select some features as common features and randomly divide the remaining features equally to both parties. Common features are not used in training for all methods except Combine. To simulate the real-world applications, for synthetic datasets, we also add different scales $\sigma_{cf}$ of Gaussian noise to the common features. Specifically, for each identifier $\mathbf{v}_i$, the perturbed identifier $\mathbf{v}_i'=\mathbf{v}_i+N(\sigma_{cf}^2\mathbf{I})$ will be used for linkage. For datasets with numeric identifiers (\texttt{house}, \texttt{taxi}, \texttt{hdb}, \texttt{syn} \texttt{boone}, \texttt{frog}), Euclidean distance is adopted to calculate similarities. For \texttt{game} dataset, Levenshtein distance is adopted to calculate similarities. For \texttt{company} dataset, we first generate Bloom filters from strings following \cite{karapiperis2017federal}, then calculate similarities based on Hamming distances.

\begin{table*}[t!]
    \centering
    \small
    \setlength\tabcolsep{3pt}
     \caption{Performance on real-world datasets}\label{tab:exp_real}
    \begin{tabular}{cccc ccc}
    \toprule
    \multirow{3}{*}{\textbf{Algorithms}} & \multicolumn{1}{c}{\textbf{house (numeric)}} & \multicolumn{1}{c}{\textbf{bike (numeric)}} & \multicolumn{1}{c}{\textbf{hdb (numeric)}} & \textbf{game (string)} & \multicolumn{1}{c}{\textbf{company (string)}} \\
    & \multicolumn{1}{c}{$\Delta=34.05$} & \multicolumn{1}{c}{$\Delta=14.26$} & \multicolumn{1}{c}{$\Delta=20.69$} & $\Delta=4.14$ & \multicolumn{1}{c}{$\Delta=10.50$}\\
    \midrule

Solo&58.31\textpm 0.28&272.83\textpm 1.50&29.75\textpm 0.15&85.27\textpm 0.29\%&42.67\textpm 0.66\\
Exact&-&-&-&89.25\textpm 0.12\%&44.44\textpm 1.95\\
Top1Sim&58.54\textpm 0.35&256.19\textpm 1.39&31.56\textpm 0.21&92.71\textpm 0.08\%&42.84\textpm 0.77\\
FeatureSim&66.39\textpm 0.15&273.29\textpm 0.37&37.39\textpm 0.29&91.13\textpm 0.23\%&39.24\textpm 1.80\\
AvgSim&51.92\textpm 0.65&239.85\textpm 0.40&34.12\textpm 0.19&90.84\textpm 0.14\%&38.19\textpm 0.91\\
\midrule
FedSim (w/o Weight)&42.82\textpm 0.20&236.79\textpm 0.29&27.18\textpm 0.08&92.79\textpm 0.13\%&41.00\textpm 1.19\\
FedSim (w/o Sort)&52.14\textpm 0.58&238.30\textpm 0.81&36.35\textpm 0.42&92.79\textpm 0.10\%&38.28\textpm 1.56\\
FedSim (w/o CNN)&42.62\textpm 0.20&235.97\textpm 0.42&27.76\textpm 0.13&92.50\textpm 0.12\%&39.63\textpm 1.31\\
\textbf{FedSim}&\textbf{42.12\textpm 0.23}&\textbf{235.67\textpm 0.27}&\textbf{27.13\textpm 0.06}&\textbf{92.88\textpm 0.11}\%&\textbf{37.08\textpm 0.61}\\

    \bottomrule
    \end{tabular} 
\end{table*}

\textbf{Training.}  Similarity model is a multi-layer perceptron (MLP) with one hidden layer. The merge model contains a 2D convolutional layer with $k_{conv}\times 1$ kernel followed by a dropout layer and an MLP with one hidden layer. Both the local model and aggregate model in SplitNN are MLPs with one hidden layer. We adopt LAMB~\cite{you2019large}, the state-of-the-art large-batch optimizer, to train all the models. Each dataset is split into training, validation, and test sets by 7:1:2. We run each algorithm five times and report the mean and standard variance (range is reported instead in figures) of performance on the test set. We present root mean square error (RMSE) or R-squared value ($R^2$) for regression tasks and accuracy for classification tasks. The choices of hyperparameters are introduced in Appendix~\ref{sec:exp_detail}.

\textbf{Baselines.}
We compare FedSim with nine baselines in our experiments. Besides the four baselines (Exact, Top1Sim, AvgSim, FeatureSim) introduced in Section~\ref{subsec:baseline}, the remaining baselines include:
1) Solo: only dataset $\mathbf{d}^P$ is trained; 
2) Combine: $[\mathbf{d}^P,\mathbf{d}^S]$ is trained (only applicable for synthetic datasets);
3) FedSim-MLP: the CNN in FedSim is changed to an MLP with a similar number of parameters.
4) FedSim w/o sort: FedSim without sorting gate.
5) FedSim w/o weight: FedSim without weight gate (similarities are directly regarded as weights).
Notably, \texttt{Exact} is only evaluated on \texttt{game} and \texttt{company} because no exactly matched identifiers can be found on other datasets.

\subsection{Performance}\label{subsec:exp_perf}

\begin{figure*}[t!]
    \centering
    \begin{subfigure}{.72\textwidth}
    \includegraphics[width=0.32\linewidth]{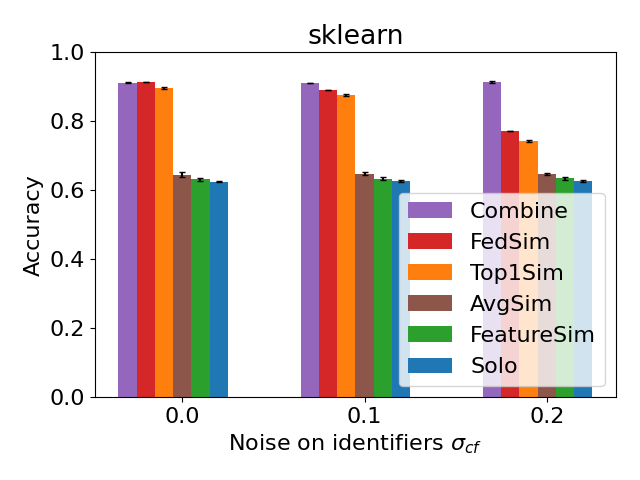}
     \includegraphics[width=0.32\linewidth]{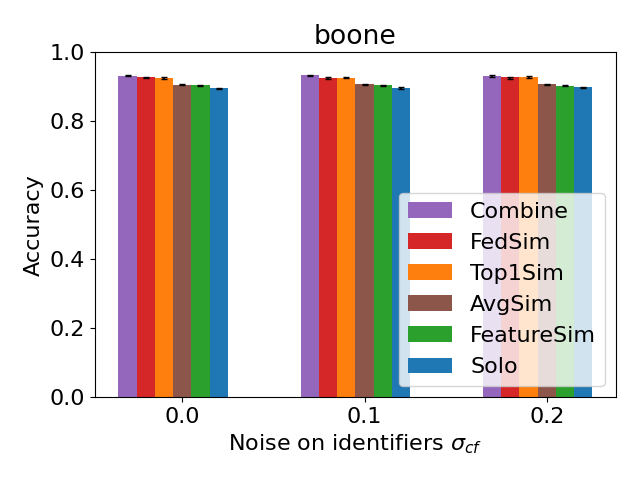}
      \includegraphics[width=0.32\linewidth]{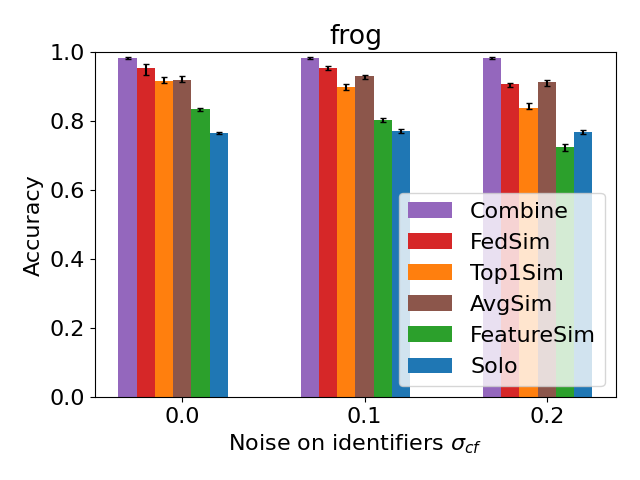}
      \caption{Performance on synthetic datasets}\label{fig:exp_euclidean_syn}
    \end{subfigure}
    \begin{subfigure}{.27\textwidth}
    \includegraphics[width=.8\linewidth]{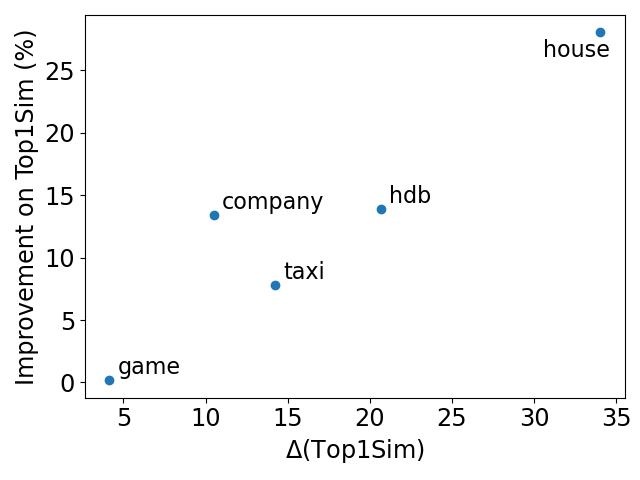}
    \caption{$\Delta(\text{Top1Sim})$ vs. perf.}\label{fig:metric_improve}
    \end{subfigure}
    \caption{Performance on synthetic datasets and the effectiveness of $\Delta(\text{Top1Sim})$}
\end{figure*}

\begin{figure*}[t!]
    \centering
    \includegraphics[width=0.24\linewidth]{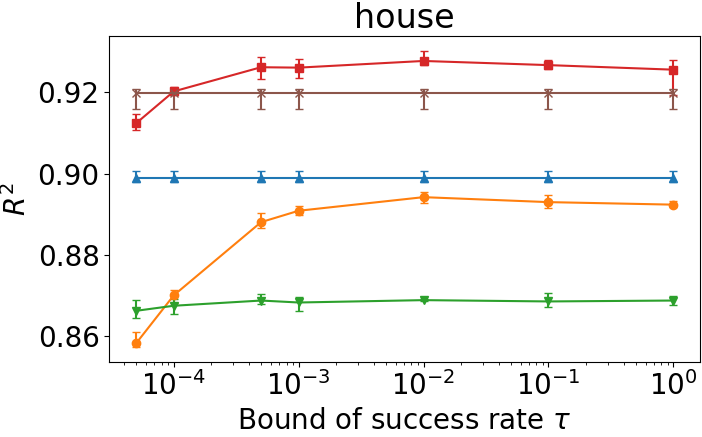}
    \includegraphics[width=0.24\linewidth]{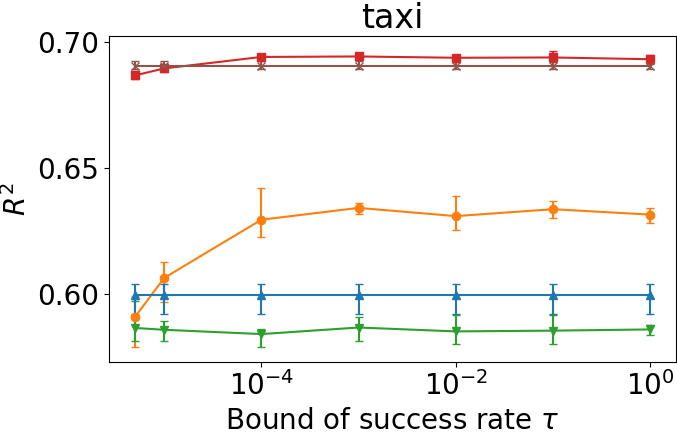}
    \includegraphics[width=0.24\linewidth]{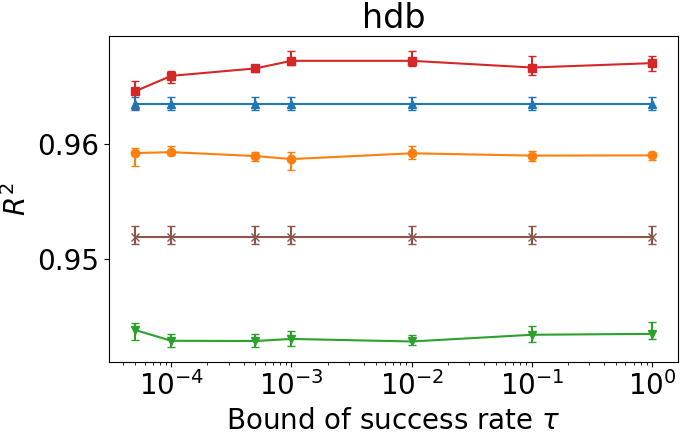}
    \includegraphics[width=0.24\linewidth]{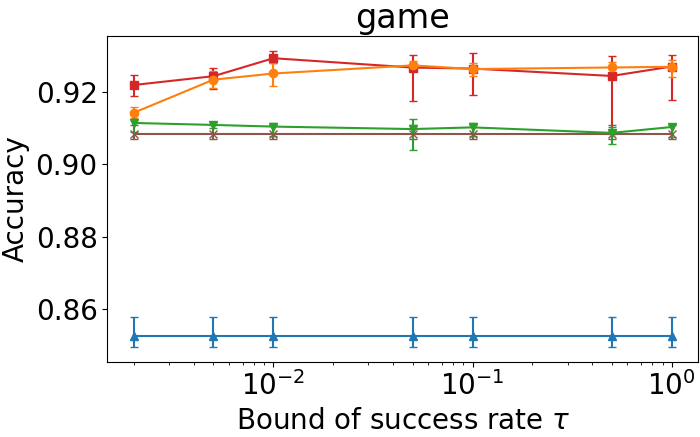}
    \includegraphics[width=0.4\linewidth]{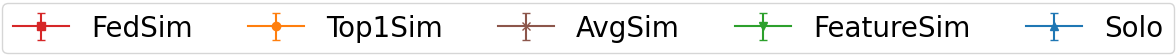}
    \caption{Performance with different scale of noise on similarities}
    \label{fig:exp_hamming_real}
\end{figure*}

We evaluate the performance of FedSim on three synthetic datasets under different $\sigma_{cf}$ and five real-world datasets. The results of synthetic datasets are presented in Figure~\ref{fig:exp_euclidean_syn}, from which two observations can be made. First, FedSim consistently has better or close performance compared to all the baselines. Second, FedSim is more robust to the noise on the identifiers. For example, in \texttt{frog}, the accuracy of Top1Sim drops to 84\% as $\sigma_{cf}=0.2$, while the accuracy of FedSim remains 91\%.

The results of real-world datasets are summarized in Table~\ref{tab:exp_real}. We also calculate the estimated improvement $\Delta(Top1Sim)$ (denoted as $\Delta$ for simplicity) according to our proposed metric. The relationship between $\Delta(Top1Sim)$ and the relative improvement on Top1Sim is presented in Figure~\ref{fig:metric_improve}. Three observations can be made from the results. First, FedSim consistently produces the best performance on all five datasets, while the baselines \wzm{(especially the separated approaches)} only have good performances on specific datasets. For example, Top1Sim has close performance to FedSim on \texttt{game}, but fails on \texttt{bike}; AvgSim has close performance to FedSim on \texttt{bike}, but fails on \texttt{game}. Second, $\Delta(Top1Sim)$ is positively correlated with real improvement on Top1Sim, indicating that the metric can be effectively used to estimate the improvement of FedSim without training. This also implies that FedSim can effectively reduce the effect of lost information and redundant information as expected. Third, comparing the performance of removing each component from FedSim, the sort gate makes the most significant contribution to the performance of FedSim by stabilizing the updates of the merge model. The improvement of the weight gate indicates that adjusting the distribution of similarities can slightly benefit the performance. Besides, CNN merge gate can also slightly improve on MLP merge gate by reducing overfitting.

\wzmc{As elaborated in Section~\ref{apdx:application}, FedSim only boosts performance on the datasets satisfying a \textbf{widely held} assumption: the similarity between identifiers is related to the similarity between data records. Our experiments in Appendix~\ref{apdx:random_feature} indicate that FedSim has close (small $K$) or lower (large $K$) performance compared to baselines on a synthetic dataset with independent identifiers. }

\subsection{Privacy}\label{subsec:exp_priv}
In this subsection, to study how additional noise on similarities affects the performance of FedSim, we conduct experiments on five real-world datasets (\wzm{the result of \texttt{company} is included in Appendix~\ref{subsec:exp_company} due to page limit}). Specifically, string or numeric identifiers are converted to Bloom filters according to \cite{karapiperis2017federal}. The Hamming distances between Bloom filters are used to calculate raw similarities. Then, given an acceptable success rate $\tau$, a noise scale $\sigma$ is calculated according to Theorem~\ref{thm:main}. Finally, Gaussian noise with scale $\sigma$ is added to the raw similarities according to Equation~\ref{eq:add_noise}. The results are presented in Figure~\ref{fig:exp_hamming_real}. \texttt{Exact} is not evaluated since few Bloom filters have exactly the same bits.
From Figure~\ref{fig:exp_hamming_real}, we observe that FedSim is robust to the noise on similarities; therefore, the attacking success rate can be reduced to $[10^{-4}, 10^{-3}]$ without evident performance loss. Notably, the performance when $\tau=1$ is not necessarily the same as the performance in Section~\ref{subsec:exp_perf} since similarities are calculated based on different distances.

\section{Related Work}\label{sec:related_work}
Most studies \cite{fu2021vf2boost,vepakomma2018split,luo2021feature,hu2019fdml,sharma2019secure,wu2022practical} in VFL focus on training and simply assumes record linkage has been done (i.e., the implicit exact linkage on record ID), which is impractical since most real-world federated datasets are unlinked. Some approaches exactly link the identifiers by exact PPRL \cite{cheng2019secureboost} or private set intersection (PSI) \cite{cheng2019secureboost,liu2020asymmetrically,wu2020privacy}. Nonetheless, these approaches incur performance loss of VFL and are also impractical since the common features of many real-world federated datasets cannot be exactly linked (e.g. GPS location). \cite{hardy2017private} greedily links the most similar identifiers in PPRL, which negatively impacts performance since some beneficial pairs with relatively low similarity may be neglected. \cite{nock2018entity,nock2021impact} explore the impact of record linkage on the performance of VFL, which is also adopted by \cite{kang2020fedmvt}. However, all of them focus only on the most similar identifiers and assume there is a one-to-one mapping between the data records of two parties, which is not always true in practice.

Current VFL frameworks support various machine learning models including linear regression \cite{feng2020multi}, logistic regression \cite{hu2019learning}, support vector machine \cite{liu2019communication}, gradient boosting decision trees \cite{cheng2019secureboost,wu2020privacy}. FDML \cite{hu2019fdml} supports neural networks but it requires all the parties to hold labels. SplitNN \cite{vepakomma2018split} focuses on neural networks and provides a new idea of collaborative learning where the model is split and held by multiple parties. Since we study the scenario where only one party holds the labels and want to support commonly used neural networks, we build FedSim on top of SplitNN.
 
\section{Conclusion}\label{sec:conclusion}
In this paper, we propose FedSim, a novel VFL framework based on similarities to boost the performance of VFL by directly utilizing the similarities calculated in PPRL and skipping the classification process. We also theoretically analyze the additional privacy risk introduced by sharing similarities and provide a bound for the success rate of an intuitive attack. In our experiment, FedSim consistently outperforms all the baselines. %

\section*{Acknowledgements}
\wzmc{This research is supported by the National Research Foundation, Singapore under its AI Singapore Programme (AISG Award No: AISG2-RP-2020-018). Any opinions, findings and conclusions or recommendations expressed in this material are those of the authors and do not reflect the views of National Research Foundation, Singapore. Qinbin is also in part supported by a Google PhD Fellowship.}

\bibliographystyle{plain}

\bibliography{references}

\section*{Checklist}

\begin{enumerate}

\item For all authors...
\begin{enumerate}
  \item Do the main claims made in the abstract and introduction accurately reflect the paper's contributions and scope?
    \answerYes{See the last paragraph of Section~\ref{sec:intro}.}
  \item Did you describe the limitations of your work?
    \answerYes{See Appendix~\ref{sec:limitation}.}
  \item Did you discuss any potential negative societal impacts of your work?
    \answerYes{See Appendix~\ref{sec:discussion}.}
  \item Have you read the ethics review guidelines and ensured that your paper conforms to them?
    \answerYes{}
\end{enumerate}

\item If you are including theoretical results...
\begin{enumerate}
  \item Did you state the full set of assumptions of all theoretical results?
    \answerYes{See Section~\ref{sec:privacy}.}
        \item Did you include complete proofs of all theoretical results?
    \answerYes{See Appendix~\ref{sec:proof}.}
\end{enumerate}

\item If you ran experiments...
\begin{enumerate}
  \item Did you include the code, data, and instructions needed to reproduce the main experimental results (either in the supplemental material or as a URL)?
    \answerYes{The codes are included in the supplementary materials. The data are all publicly available; the link or citations of data are presented in Appendix~\ref{sec:exp_detail}.}
  \item Did you specify all the training details (e.g., data splits, hyperparameters, how they were chosen)?
    \answerYes{All the details including data splits and hyperparameters are introduced in Section~\ref{subsec:exp_setup} and Appendix~\ref{sec:exp_detail}.}
        \item Did you report error bars (e.g., with respect to the random seed after running experiments multiple times)?
    \answerYes{The variance is reported in all tables and figures.}
        \item Did you include the total amount of compute and the type of resources used (e.g., type of GPUs, internal cluster, or cloud provider)?
    \answerYes{See the ``hardware'' paragraph of Appendix~\ref{sec:exp_detail}.}
\end{enumerate}

\item If you are using existing assets (e.g., code, data, models) or curating/releasing new assets...
\begin{enumerate}
  \item If your work uses existing assets, did you cite the creators?
    \answerYes{See Section~\ref{subsec:exp_setup} and Appendix~\ref{sec:exp_detail}.}
  \item Did you mention the license of the assets?
    \answerYes{See the ``license'' paragraph of Appendix~\ref{sec:exp_detail}.}
  \item Did you include any new assets either in the supplemental material or as a URL?
    \answerYes{Our codes are included in supplementary materials.}
  \item Did you discuss whether and how consent was obtained from people whose data you're using/curating?
    \answerYes{See the ``Consent of dataset'' paragraph of Appendix~\ref{sec:discussion}}
  \item Did you discuss whether the data you are using/curating contains personally identifiable information or offensive content?
    \answerYes{See the ``Personally identifiable information'' of Appendix~\ref{sec:discussion}.}
\end{enumerate}

\item If you used crowdsourcing or conducted research with human subjects...
\begin{enumerate}
  \item Did you include the full text of instructions given to participants and screenshots, if applicable?
    \answerNA{Our research is not related to human subjects.}
  \item Did you describe any potential participant risks, with links to Institutional Review Board (IRB) approvals, if applicable?
    \answerNA{Our research is not related to human subjects.}
  \item Did you include the estimated hourly wage paid to participants and the total amount spent on participant compensation?
    \answerNA{Our research is not related to human subjects.}
\end{enumerate}

\end{enumerate}

\clearpage
\appendix
\section{Proofs and Analysis of Privacy}\label{sec:proof}

\subsection{Proof of Differential Privacy (DP)} \label{subsec:proof_dp}

The privacy risk of FedSim incurred by similarities can also be formalized by a strong metric \textit{differential privacy}~\cite{dwork2014algorithmic}. This metric, however, requires more additional noise, thus seriously affecting the accuracy. In the beginning, we introduce some background of traditional Gaussian mechanism (Theorem~\ref{thm:gaussian}) \cite{dwork2014algorithmic} as $\varepsilon\in(0,1)$.

\begin{theorem}\label{thm:gaussian} (\cite{dwork2014algorithmic})
    For any $(\varepsilon,\delta)\in (0,1)$, the Gaussian mechanism $M(x)=f(x)+N(0,\sigma^2I)$ with $\sigma=\Delta \sqrt{2\log(1.25/\delta}$.
\end{theorem}

According to \cite{balle2018improving}, the traditional Gaussian mechanism can be extended to Theorem~\ref{thm:gaussian_adv} to support $\varepsilon>1$.

\begin{theorem}\label{thm:gaussian_adv} (\cite{balle2018improving})
    Let $f:\mathbb{X}\rightarrow \mathbb{R}^d$ have global $L_2$ sensitivity $\Delta$. Suppose $\varepsilon > 0$ and $0 < \delta <1/2-e^{-3\varepsilon}/\sqrt{2\pi\varepsilon}$. If the mechanism $M(x)=f(x)+N(0,\sigma^2I)$ is $(\varepsilon,\delta)$-DP, then $\sigma\ge \Delta/\sqrt{2\varepsilon}$.
\end{theorem}

Theorem~\ref{thm:gaussian} suggests that the Gaussian mechanism as $\varepsilon>1$ (Theorem~\ref{thm:gaussian}) requires different scale of noise compared to the Gaussian mechanism as $\varepsilon\in(0,1)$ (Theorem~\ref{thm:gaussian_adv}). Specifically, to achieve $(\varepsilon,\delta)$-DP, Theorem~\ref{thm:gaussian} requires $\sigma=\Uptheta(1/\varepsilon)$, while Theorem~\ref{thm:gaussian_adv} requires $\sigma=\Omega(1/\sqrt{\varepsilon})$. Notably, Theorem~\ref{thm:gaussian_adv} indicates the smallest scale $\sigma$ of noise required for $(\varepsilon,\delta)$-DP, thus \textbf{further implying the lowest $\varepsilon$ that can be proved from the given scale of noise ($\sigma\ge \Delta/\sqrt{2\varepsilon}$)}. Focusing on the case of FedSim, given the dataset $\mathbf{k}^S$ to be protected, we restate procedure $\mathcal{G}$ as follows.

Then, we prove the differential privacy of $\mathcal{G}$ in both cases when $\varepsilon\in(0,1)$ 
and $\varepsilon>1$ according to Gaussian mechanisms.

\paragraph{Procedure $\mathcal{G}$:} Take $\mathbf{k}^S$ as the input. Each $k_i^P\in \mathbf{k}^P$ is linked to multiple $k_j^S\in \mathbf{k}^S$; the similarities between $k_i^P$ and $k_j^S$ are calculated by $s_{ij}=-\frac{\|k_i^P-k_j^S\|_2+\mu_0}{\sigma_0}+N(0,\sigma^2)$, where $N(0,\sigma^2)$ refers to Gaussian distribution with variance $\sigma^2$. Output the similarities $\mathbf{s}$.

\begin{theorem}\label{thm:dp_sim}
Let $\varepsilon,\delta\in(0,1)$. Suppose the size of $\mathbf{k}^P$ is $n\times \beta$. Procedure $\mathcal{G}$ is $(\varepsilon,\delta)$-DP if $\sigma\ge \Delta_\mathcal{G}\sqrt{2\ln{(1.25/\delta})}/\varepsilon$, where $\Delta_\mathcal{G}=n\cdot \max\left\{\left|\frac{1+\mu_0}{\sigma_0}\right|,\left|\frac{-1+\mu_0}{\sigma_0}\right|\right\}$. 
\end{theorem}
\begin{proof}
The change of a single of one Bloom filter on party $S$ may change the distance by 1; therefore, the sensitivity of $\|k_i^P-k_j^S\|_2=1$. The sensitivity of a single similarity $s_{ij}$, which is a linear transformation of $\|k_i^P-k_j^S\|_2$, is 
\begin{equation}
    \max\left\{\left|\frac{1+\mu_0}{\sigma_0}\right|,\left|\frac{-1+\mu_0}{\sigma_0}\right|\right\}
\end{equation}
Since each $k_j^S\in \mathbf{k}^S$ might be linked to all $n$ instances in $\mathbf{k}^P$, modifying one instance in $\mathbf{k}^S$ results in the change of at most $n$ similarities. The sensitivity of procedure $\mathcal{G}$ can be derived as
\begin{equation}\label{eq:sensitivity}
    \Delta_\mathcal{G}= n\cdot \max\left\{\left|\frac{1+\mu_0}{\sigma_0}\right|,\left|\frac{-1+\mu_0}{\sigma_0}\right|\right\}
\end{equation}
According to Theorem~\ref{thm:gaussian}, if $\sigma\ge \Delta_\mathcal{G}\sqrt{2\ln{(1.25/\delta})}/\varepsilon$, procedure $\mathcal{G}$ is differential privacy $(\varepsilon,\delta)$-differential privacy according to Gaussian mechanism.
\end{proof}

\dpFedSimAdv*
\begin{proof}
    According to Equation~\ref{eq:sensitivity}, it holds
    \[
        \Delta_\mathcal{G}= n\cdot \max\left\{\left|\frac{1+\mu_0}{\sigma_0}\right|,\left|\frac{-1+\mu_0}{\sigma_0}\right|\right\}
    \]
    According to Theorem~\ref{thm:gaussian_adv}, if $\mathcal{G}$ is $(\varepsilon,\delta)$-DP, then $\sigma\ge \Delta_\mathcal{G}/\sqrt{2\varepsilon}$.
\end{proof}

\subsection{Proof of Theorem~\ref{alg:main}}\label{subsec:proof_main}
\main*

\begin{proof}
	According to the attack method, the attacker can predict based on any \(|Q|\) Bloom filters in party P and their corresponding distances. We denote the negative distance between $k^P_i$ and $k^S_j$ as \(l_{ij}=-\dist(k^P_i,k^S_j)\). The attacker's predicted value of \(l_{ij}\) is denoted as \(\hat{l}_{ij}\). Then, we have
	\begin{equation}\label{eq:main_prob_init}
	\small
		\begin{aligned}
			&\;\Pr\left[\left.\hat{k}^S_j=k^S_j\right|\{s_{ij}|i\in Q\},\{k^P_{i}|i\in Q\},\mu_0,\sigma_0,\mathcal{A}\right] \\
			=&\;\Pr\left[\left.\hat{k}^S_j=k^S_j,\{\hat{l}_{ij}=l_{ij}|i\in Q\}\right|\{s_{ij}|i\in Q\},\{k^P_{i}|i\in Q\},\mu_0,\sigma_0,\mathcal{A}\right] \\
			=&\;\Pr\left[\left.\hat{k}^S_j=k^S_j\right|\{\hat{l}_{ij}=l_{ij}|i\in Q\},\{k^P_{i}|i\in Q\},\mu_0,\sigma_0,\mathcal{A}\right]\\
			&\;\cdot\Pr\left[\left.\{\hat{l}_{ij}=l_{ij}|i\in Q\}\right|\{s_{ij}|i\in Q\},\{k^P_{i}|i\in Q\},\mu_0,\sigma_0,\mathcal{A}\right]\\
			=&\;\Pr\left[\left.\hat{k}^S_j=k^S_j\right|\{\hat{l}_{ij}=l_{ij}|i\in Q\},\{k^P_{i}|i\in Q\},\mu_0,\sigma_0,\mathcal{A}\right] \\
			&\cdot \Pr\left[\left.\{\hat{l}_{ij}=l_{ij}|i\in Q\}\right|\{s_{ij}|i\in Q\},\mathcal{A}\right] \\
			=&\;\Pr\left[\left.\hat{k}^S_j=k^S_j\right|\{\hat{l}_{ij}=l_{ij}|i\in Q\},\{k^P_{i}|i\in Q\},\mu_0,\sigma_0,\mathcal{A}\right]  \\ &\cdot\prod_{i\in Q}\Pr\left[\left.\hat{l}_{ij}=l_{ij}\right|s_{ij},\mathcal{A}\right]
		\end{aligned}
	\end{equation}
	The first equation holds because of the assumption that the attacker predicts \(k^S_j\) through the most likely distance $\hat{l}_{ij}$, and correctly predicting the $|Q|$ correct distances is the prerequisite of predicting the correct Bloom filter \(k^S_j\), i.e., \(\{\hat{k}^S_j=b^S_j\}\subseteq\{\hat{l}_{ij}=l_{ij}|i\in Q\}\). The second equation holds because of the definition of conditional probability. The third equation holds because the attacker predicts distances independently based on perturbed similarities, indicating that $\{\hat{l}_{ij}=l_{ij}|i\in Q\}$ is \textbf{conditionally independent} with $\mu_0,\sigma_0$ and $\{k^P_{i}|i\in Q\}$ given $\mathcal{A}$. The fourth equation holds because all the \(\hat{l}_{ij}\;(i\in Q)\) are predicted independently by the attacker from the corresponding \(s_{ij}\), indicating that each $\hat{l}_{ij}=l_{ij}$ is \textbf{conditionally independent} with $\{s_{kj}|k\in Q,k\neq i\}$.
	
	According to the attack method, the attacker first predicts the similarity \(\hat{\rho}_{ij}\) based on Maximum a Posteriori (MAP) estimation with one experiment \(s_{ij}\). With Bayes' theorem, we have
	\begin{equation}
		 p(\rho_{ij}|s_{ij},\mathcal{A})\propto p(s_{ij}|\rho_{ij},\mathcal{A})p(\rho_{ij}|\mathcal{A})
	\end{equation}
	Since \(\rho_{ij}\sim N(0,1)\) and \(s_{ij}\sim N(\rho_{ij},\sigma^2)\),
	\begin{equation}\label{eq:post_dist}
		 p(\rho_{ij}|s_{ij},\mathcal{A})\sim N\left(\frac{s_{ij}}{\sigma^2+1},\frac{\sigma^2}{\sigma^2+1}\right)
	\end{equation}
	Note that the posterior distribution is also a Gaussian distribution. This property of Gaussian distribution is also known as \textit{conjugate distribution} \cite{murphy2007conjugate}. With Equation~\ref{eq:post_dist}, \(\rho_{ij}\) is estimated by
	\begin{equation}\label{eq:map}
		\hat{\rho_{ij}}=\argmax_{\rho_{ij}} p(\rho_{ij}|s_{ij},\mathcal{A})=\frac{s_{ij}}{\sigma^2+1}
	\end{equation}
	Then, \(\hat{\rho_{ij}}\) is scaled back to distance $\hat{l_{ij}}'$ with known scaling parameters \(\mu_0,\sigma_0\), i.e.,
	\begin{equation}\label{eq:d_i_hat_prime}
		\begin{aligned}
			\hat{l_{ij}}' = \sigma_0\hat{\rho_{ij}}+\mu_0
		\end{aligned}
	\end{equation}
	Since the distances between the Bloom filters are integers, predicting the correct distance, i.e., \(\hat{l}_{ij}=l_{ij}\), implies that
	\begin{equation}\label{eq:round_dist}
	\begin{aligned}
	    {l_{ij}} = \sigma_0{\rho_{ij}}+\mu_0,  |\hat{l_{ij}}'-l_{ij}|\le \frac{1}{2}
	\end{aligned}
	\end{equation}
	With Equation \ref{eq:map}, \ref{eq:d_i_hat_prime} and \ref{eq:round_dist}, we can evaluate \(\Pr[\hat{l}_{ij}=l_{ij}|s_{ij},\mathcal{A}]\) by 
	\begin{equation}\label{eq:single_d_i_prob}
\begin{aligned}
\Pr[\hat{l}_{ij}=l_{ij}|s_{ij}]&=\Pr\left[\left.|\hat{l_{ij}}'-l_{ij}|\le \frac{1}{2}\right|s_{ij},\mathcal{A}\right] \\ &=\Pr\left[\left.\left|\rho_{ij}-\frac{s_{ij}}{\sigma^2+1}\right|\le \frac{1}{2\sigma_0}\right|s_{ij},\mathcal{A}\right]
\end{aligned}
	\end{equation}
	Note that we already know the probability density function \(p(\rho_{ij}|s_{ij},\mathcal{A})\) in Equation~\ref{eq:post_dist}. By shifting the mean of the distribution, we have
	\begin{equation}\label{eq:shift_dist}
		p\left(\left.\rho_{ij}-\frac{s_{ij}}{\sigma^2+1}\right|s_{ij},\mathcal{A}\right)\sim N\left(0,\frac{\sigma^2}{\sigma^2+1}\right)
	\end{equation}
	Considering Equation~\ref{eq:single_d_i_prob} and \ref{eq:shift_dist}, \(\Pr[\hat{l}_{ij}=l_{ij}|s_{ij}]\) can be calculated by a simple integral
	\begin{equation}\label{eq:integral}
	\begin{aligned}
			\Pr[\hat{l}_{ij}=l_{ij}|s_{ij},\mathcal{A}] &=
		\int_{\frac{1}{-2\sigma_0}}^\frac{1}{2\sigma_0} \frac{\sqrt{\sigma^2+1}}{\sqrt{2\pi}\sigma}e^{-\frac{x^2}{2\sigma^2}(\sigma^2+1)}dx =\erf\left(\frac{\sqrt{\sigma^2+1}}{2\sqrt{2}\sigma\sigma_0}\right)
	\end{aligned}
	\end{equation}
From Equation~\ref{eq:main_prob_init}, \ref{eq:single_d_i_prob}, and \ref{eq:integral}, we have
\begin{equation}\label{eq:all_prob_bound}
	\begin{aligned}
	&\;\Pr\left[\left.\hat{k}^S_j=k^S_j\right|\{s_{ij}|i\in Q\},\{k^P_{i}|i\in Q\},\mu_0,\sigma_0,\mathcal{A}\right] \\
	=&\;\Pr\left[\left.\hat{k}^S_j=k^S_j\right|\{\hat{l}_{ij}=l_{ij}|i\in Q\},\{k^P_{i}|i\in Q\},\mu_0,\sigma_0,\mathcal{A}\right] \\
	&\cdot\prod_{i\in Q} \Pr\left[\left.\hat{l}_{ij}=l_{ij}\right|s_{ij},\mathcal{A}\right] \\
	\le&\;\prod_{i\in Q}\Pr[\hat{l}_{ij}=l_{ij}|s_{ij},\mathcal{A}]=\;\left[\erf\left(\frac{\sqrt{\sigma^2+1}}{2\sqrt{2}\sigma\sigma_0}\right)\right]^{|Q|}
	\end{aligned}
\end{equation}
The inequality in the fourth line of Equation~\ref{eq:all_prob_bound} holds because the first probability is always less or equal to 1. This relaxation implies that the properties of Bloom filters (e.g., size, value) does not affect the proved bound.

According to the property of error function, \(0<\erf(x)<1\) for any \(x>0\). Thus, for any set \(Q\), we have
\begin{equation}\label{eq:erf_k}
\begin{aligned}
        		&\Pr\left[\left.\hat{k}^S_j=k^S_j\right|\{s_{ij}|i\in Q\},\{k^P_{i}|i\in Q\},\mu_0,\sigma_0,\mathcal{A}\right] \le \erf\left(\frac{\sqrt{\sigma^2+1}}{2\sqrt{2}\sigma\sigma_0}\right)
\end{aligned}
\end{equation}

\end{proof}

Interestingly, Equation~\ref{eq:all_prob_bound} implies that a larger $|Q|$ leads to a lower success rate of the attack, which is reasonable due to the noise added to each $s_{ij}$. Specifically, because of the added noise in Equation~\ref{eq:add_noise}, the attacker cannot accurately predict the real similarities in the first step of the attack. Though the attacker has more predicted real similarities $\hat{\rho_i}$ to infer $k_j^S$ in the last step of the attack, these additional $\hat{\rho_i}$ bring more noise than information, thus leading to a higher success rate. In conclusion, under the intuitive greedy attack model, the optimal choice of the attacker is to launch an attack through a single Bloom filter and its similarity by setting $|Q|=1$, the success rate of which can be bounded by a small constant.

\subsection{Comparison between Two Privacy Metrics}
\begin{figure}[htpb]
    \centering
    \includegraphics[width=0.4\textwidth]{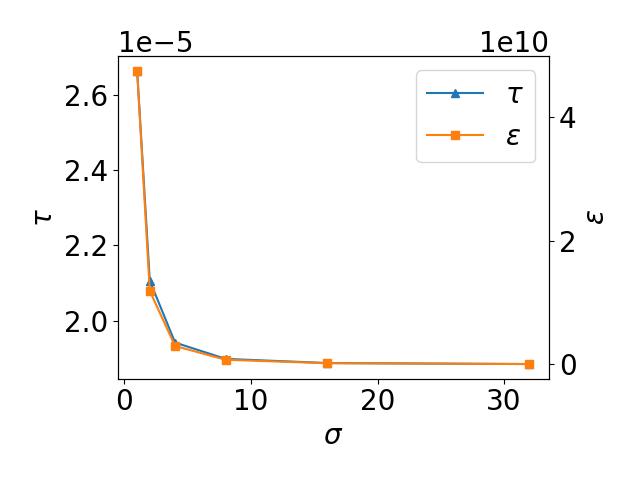}
    \caption{Values of $\tau$ and $\varepsilon$ under different noise scales $\sigma$}
    \label{fig:dp_tau}
\end{figure}

As observed from both Theorem~\ref{thm:dp_sim} and Theorem~\ref{thm:dp_sim_adv}, the noise scale $\sigma$ derived from differential privacy increases linearly by $n$. A such large scale of noise would seriously affect the performance of FedSim. Taking \texttt{house} dataset as an example, $\mu_0=-46237.78, \sigma_0=21178.86$. Letting $\delta=10^{-5}$, we can derive the values of $\tau$ and $\varepsilon$ under different noise scales $\sigma$ as Figure~\ref{fig:dp_tau}. As observed from Figure~\ref{fig:dp_tau}, setting $\sigma=4$, the differential privacy parameter $\varepsilon=2.96\times 10^9$ implies that there is almost no privacy guarantee at all. Nonetheless, the attacking success rate $\tau=1.94\times 10^{-5}$ suggests that the privacy risk from certain attacks can be very low.

The main difference between the two metrics is the accumulation of privacy risk over various similarities linked to the same $k_j^S$. In differential privacy, the overall privacy loss is a linear summation of the privacy loss from each similarity. Nevertheless, in our proposed metric, the attacking success rate $\tau$ decreases as the number $|Q|$ of similarities used in the attack increases according to Equation~\ref{eq:all_prob_bound}. In summary, \textbf{differential privacy that protects against all possible attacks is impractical in this scenario; as an alternative, our analysis suggests that the privacy risk against certain attacks can be significantly reduced by adding Gaussian noise}.

\section{Advanced Attacks on FedSim}\label{sec:adv_attack}
\subsection{Advanced Attacks}

\rev{Since FedSim is the first VFL approach indicating that the proper use of similarities can benefit the training process, the attacks on the similarities remain unexplored. In this paper, we have proposed a greedy attack method and the corresponding defense as a start. Notably, more advanced attacks are non-trivial but possible and their corresponding defense mechanisms are desirable. We provide three potential vulnerabilities of FedSim for future attack design.}

\textbf{1) Background information of $\rho_{ij}$:} \rev{In step (1) of the greedy attack method, we assume the attack adopts a Gaussian prior distribution to perform estimation, which is true when the attacker has no knowledge of the distribution of distances $\rho_{ij}$. Nonetheless, if some prior knowledge of $\rho_{ij}$ is known to the attacker, one may carefully design a better prior than unit Gaussian and achieve higher predictive accuracy.
For example, if knowing $\rho_{ij} < 100$, the attacker can truncate the prior distribution by setting $\Pr[\rho_{ij}\ge 100]=0$.}
\textbf{2) Non-greedy attack:} The greedy attack method assumes the attacker predicts Bloom filters with the predicted values of distances. However, deriving an accurate value of each distance may not be necessary. It remains an open direction to design a method that directly exploits the distribution of distances to predict the target Bloom filter.
\textbf{3) Correlation between shared information:} In  Section~\ref{subsec:priv_analyze}, we list three shared information and discuss their privacy risk respectively. Nevertheless, the privacy of each piece of shared information does not imply the privacy of all three pieces of shared information. For example, if shared similarities $\mathbf{s}$ are correlated with shared intermediate results ${\mathbf{c}_i}_{i=1}^m$ in SplitNN, disclosing both could much more damaging than disclosing each of them. The effect of such correlation, which remains unexplored, will be left as our future work.

\section{Extension to Multiple Parties}\label{sec:multi_party}

As the effect of linkage on VFL receives increasing attention \cite{nock2021impact}, to the best of our knowledge, all the existing VFL approaches with non-exact linkage \cite{nock2018entity,kang2020fedmvt,hardy2017private,nock2021impact} focus on the two-party setting. Similar to these approaches, we also mainly focus on the two-party setting which has many real-world applications (e.g., bank and fintech company \cite{wu2020privacy}). In this section, we explain the major challenges of multi-party VFL with non-exact linkage and present a simple extension of FedSim to the multi-party setting.

The multi-party setting of VFL is far more than a trivial extension of the two-party setting because the linkage among parties can be complex. For example, consider a three-party VFL of an e-bank, an e-shopping company, and a delivery company. The e-bank and e-shopping company are linked by customers' \textit{names}; the e-shopping company and the delivery company are linked by \textit{transaction IDs}; the e-bank and the delivery company are linked by \textit{address}. In this case, we need to select two ``best'' common features from \textit{name}, \textit{transaction ID}, and \textit{address} to link the three parties and determine a proper order for the linkage. Although in the exact linkage on the same common feature \cite{cheng2019secureboost,wu2020privacy,hu2019fdml}, the choice and order of common features do not affect the linkage result, the result of non-exact linkage can be significantly affected by both the choice and order of common features. How to handle these data non-exactly linked as a circle (or even a complicated graph) is still an open problem of VFL.

Additionally, even assuming that all the parties are linked on the same common feature and the order of linkage is fixed, performing one-to-many linkage between each pair of parties suffers an efficiency issue. Specifically, one intuitive approach is to link all possible pairs across the parties and expand $\mathbf{o}_i$ from $K\times l_m$ to size $K^{n-1}\times l_m$, where $n$ is the number of parties (both primary and secondary). This approach, leading to $K^{n-2}$ times more training cost than the cost of the two-party FedSim, is not scalable on parties.

Fortunately, our empirical study on five real-world datasets suggests that, \textbf{compared to secondary parties, the primary party who owns labels usually also holds more important or at least comparable features}. Specifically, we first link each data record in the primary party with the most similar data record in the secondary party (similar to Top1Sim). Then, each data record in both the primary party and the secondary party has a label. Finally, we train the same model on the primary party and the secondary party, respectively, and summarize the performance in Table~\ref{tab:exp_solo}. Compared with the secondary party, we observe that the primary party has much better performance (thus much more important features) on \texttt{house}, \texttt{hdb}, and \texttt{song}. Meanwhile, the primary party has slightly lower but comparable performance (thus comparable features) on \texttt{bike} and \texttt{game}.

\begin{table}[htpb]
    \small
    \caption{Performance on a single party (either primary or secondary) on real-world datasets}\label{tab:exp_solo}
    \begin{center}
        \begin{tabular}{cccccc}
    \toprule
    \textbf{Dataset} & house (RMSE) & bike (RMSE) & hdb (RMSE) & game (Accuracy) & company (RMSE) \\\midrule
    \textbf{Primary} & 58.31\textpm 0.28&272.83\textpm 1.50&29.75\textpm 0.15&85.27\textpm 0.29\%&37.08\textpm 0.61\\
\textbf{Secondary} & 150.27\textpm 0.17&265.79\textpm 0.06&134.82\textpm 0.10&88.21\textpm 0.06\%&225.07\textpm 0.02\\
\textbf{Relative diff.}\textsuperscript{1} & \textbf{61.20}\%&-2.65\%&\textbf{77.94}\%&-3.34\%&\textbf{83.53}\% \\
    \bottomrule
    \end{tabular}
    \end{center}
    \textsuperscript{1} Positive means primary outperforms secondary; negative means secondary outperforms primary. The value means relative difference, i.e., $|\text{primary} - \text{secondary}| / \text{secondary}$. 
\end{table}

This observation implies that the linkage between the primary party and the other parties is the most vital in practice. Under this observation, to overcome the scalability issue of multi-party FedSim, we propose an intuitive approach that extends FedSim to the multi-party setting. Specifically, we first perform one-to-one linkage for all the secondary parties, then perform one-to-many linkage between the linked secondary parties and the primary party. During the training process, each secondary party holding a local model performs training according to multi-party SplitNN. This approach is empirically evaluated in Appendix~\ref{subsec:exp_multi}.

\section{Applications}\label{apdx:application}
\wzmb{In this section, we discuss what datasets FedSim performs well on and how common such datasets exist in real-world applications. As we have explained in Appendix~\ref{apdx:random_feature}, the main assumption of FedSim is}

\begin{center}
    \textit{the similarity between identifiers is related to the similarity between data records.}
\end{center}

\wzmb{If this assumption holds, FedSim probably outperforms Exact/Top1Sim. The potential improvement can be estimated by our proposed metric $\Delta$. If this assumption does not hold, e.g., the identifier is unique ID, the experiment in Appendix~\ref{apdx:random_feature} indicates that FedSim has the close performance to Exact/Top1Sim under a small $K$. Setting a large $K$ is likely to significantly reduce the performance due to overfitting.}

\wzmb{This assumption commonly holds in real-world applications. Since VFL is a rather recent paradigm with fewer real applications, we investigate the applications of a traditional area "record linkage" which has many existing applications. The application scope of record linkage reflects the application scope of VFL, because all the VFL algorithms require the data to be linked (either on ID or other features) before training.}

\wzmb{German Record Linkage Center (GRLC) published an article \cite{AntoniSchnell2019summary} to summarize its completed linkage projects since 2011. We summarize the information of identifiers in Table~\ref{tab:application}. }

\begin{table}[htpb]
    \centering
    \caption{The identifiers of the completed projects in GRLC since 2011 (excluding two papers written in German that we are unable to translate)}
    \label{tab:application}
    \small
    \begin{tabular}{cccc}
    \toprule
        \textbf{Application} & \textbf{Reference} & \textbf{Identifier} & \textbf{Satisfy the Assumption} \\
        \midrule
        \multirow{6}{*}{Employment} 
                & \cite{antoni2014pass} & name, birthday, street, house number, etc. & \cmark \\
                & \cite{eberle2016record} & address & \cmark \\
                & \cite{perry2016research} & address & \cmark \\
                & \cite{schild2014linking} & name, address & \cmark \\
                & \cite{coppola2013saving} & ID & \xmark \\
                & \cite{weinhardt2017linked} & ID & \xmark \\
        \midrule
        \multirow{2}{*}{Commercial} 
            & \cite{schild2016linking} & name, address & \cmark \\
            & \cite{antoni2019private} & company name & \cmark \\
        \midrule
        Migration & \cite{kroh20152013} & gender, age, address, municipality, etc. & \cmark \\
        \midrule
        Medical & \cite{gramlich2014strokes} & name, date of birth & \cmark \\
        \midrule
        Education & \cite{fuss2016unique} & ID & \xmark \\
        \midrule

        \multicolumn{3}{c}{\textbf{Number of applications satisfying the assumption}} & \textbf{8/11 (72.7\%)} \\ 
        \bottomrule
    \end{tabular}
\end{table}

\wzmb{Two observations can be made from these projects. First, among the 11 completed projects, eight projects require linking datasets without user ID. This implies that, in the majority (around 8/11) of real applications, there does not exist a shared user ID. Second, the similarity of some fields can reflect the similarity of the property, such as address (5/11 projects) in GPS or string format. Even for the field whose own similarity does not reflect record property, such as name, birth date, and postal code, the similarity of the quasi-identifier containing these fields (2/11 projects) can reflect the property. This is because records with more matched fields are more likely to belong to the same user, especially when considering typos that widely exist in practice. Therefore, the similarity of shared features is related to the property of records in many (around 8/11) real-world cases, which supports our main assumption.}

\section{Experimental Details}\label{sec:exp_detail}

\paragraph{Dataset.} The basic information of these datasets is summarized in Table~\ref{tab:dataset}. Specifically, in \texttt{{house}} dataset, party P contains housing data in Beijing collected from \textit{lianjia} \cite{house}, and party S contains renting data in Beijing collected from \textit{Airbnb} \cite{airbnb}. Two parties are linked by longitude and latitude and the task is to predict the housing price. 
In \texttt{{taxi}} dataset, party P contains taxi trajectory data in New York from TLC \cite{taxi}, and party S contains bike trajectory data in New York from Citi Bike \cite{bike}. Two parties are linked by the longitude and latitude of the source and destination of the trajectory and the task is to predict the time of the trip along this trajectory. 
In \texttt{{hdb}} dataset, party P contains HDB resale data in Singapore collected from Housing and Development Board \cite{hdb}, and party S contains recent rankings and locations of secondary schools in Singapore collected from salary.sg \cite{school}. Two parties are linked by longitude and latitude and the task is to predict the HDB resale prices. 
In \texttt{{game}} dataset, party P contains game data (e.g., prices) and the number of owners (\#owners) from Steam \cite{steam}. party S contains game ratings and comments from RAWG \cite{rawg}. Two parties are linked by game titles and the task is to classify games as popular (\#owners $>$ 20000) or unpopular (\#owners $\le$ 20000) based on ratings, prices, etc.
In \texttt{{song}} dataset, party P contains timbre values of songs extracted from \textit{million song dataset} \cite{bertin2011million}, and party S contains the basic information of songs extracted from FMA dataset \cite{defferrard2016fma}. Two parties are linked by the titles of songs and the task is to predict the publication years of songs.

\begin{table}[htpb]
    \begin{center}
    \small
    \caption{Basic information of datasets}\label{tab:dataset}
    \begin{tabular}{c c ccc ccc cc c}
    \toprule
        \multirow{2}{*}{\textbf{Dataset}} & \multirow{2}{*}{\textbf{Type}} & \multicolumn{3}{c}{\textbf{Party P}} & \multicolumn{3}{c}{\textbf{Party S}} & \multicolumn{2}{c}{\textbf{Identifiers}} & \multirow{2}{*}{\textbf{Task}} \\
        \cmidrule(lr){3-5} \cmidrule(lr){6-8} \cmidrule(lr){9-10}
        & & \#samples & \#ft. & ref & \#samples & \#ft. & ref & \#dims & type \\
        \midrule
        sklearn & Syn & 60,000 & 50 & \cite{sklearn} & 60,000 & 50 & \cite{sklearn} & 5 & float & bin-cls \\
        frog & Syn & 7,195 & 19 & \cite{frog} & 7,195 & 19 & \cite{frog} & 16 &  float & multi-cls \\
        boone & Syn & 130,064 & 40 & \cite{boone} & 130,064 & 40 & \cite{boone} & 30 & float & bin-cls \\
        house & Real & 141,050 & 55 & \cite{house} & 27,827 & 25 & \cite{airbnb} & 2 & float & reg \\
        taxi & Real & 200,000 & 964 & \cite{taxi} & 100,000 & 6 & \cite{bike} & 4 & float & reg \\
        hdb & Real & 92,095 & 70 & \cite{hdb} & 165 & 10 & \cite{school} & 2 & float & reg \\
        game & Real & 26,987 & 38 & \cite{steam} & 439,999 & 86 & \cite{rawg} & 1 & string & bin-cls \\
        company & Real &  77,225 & 91 & \cite{loan} & 220,583 & 157 & \cite{company} & 1 & string & reg \\
        \bottomrule
    \end{tabular}
    \end{center}
    \textbf{Note}: ``\#ft.'' means number of features; ``Syn'' means synthetic; ``Real'' means real-world; ``bin-cls'' means binary classification; ``multi-cls'' means multi-class classification; ``reg'' means regression.
\end{table}

These eight datasets have a wide variety in multiple dimensions. 1) \textbf{Matching of common features}: Due to the property of common features, some datasets can be exactly matched (e.g., \texttt{syn}, \texttt{frog}, \texttt{boone} without noise), some datasets can partially exactly matched (e.g, \texttt{game}, \texttt{song}), and some datasets can only be soft matched (e.g., \texttt{house}, \texttt{taxi}, \texttt{hdb}). 2) \textbf{Similarity metric}: These datasets cover three commonly used similarity metrics including Euclidean-based similarity, Levenshtein-based similarity, and Hamming-based similarity. 3) \textbf{Task}: The selected datasets cover three common tasks including binary classification, multi-class classification, and regression.

\paragraph{Hyperparameters.} The number $K$ of neighbors is chosen from $\{50,100\}$. The sizes of hidden layers are chosen from $\{100, 200, 400\}$ and remain consistent across all VFL algorithms. The learning rate is chosen from $\{3\times 10^{-4}, 10^{-3}, 3\times 10^{-3}, 1\times 10^{-2}\}$ and weight decay is chosen from $\{10^{-5},10^{-4}\}$. Since FedSim and AvgSim process $K$ times more samples per batch compared to other algorithms, their batch sizes are $1/K$ times smaller than the others. The batch size of FedSim and AvgSim are chosen from $\{32,128\}$, while the batch sizes of other algorithms are set to 4096. The training is stopped at the best performance on the validation set. 

\paragraph{Hardwares.} The training of all the experiments is conducted on a machine with four A100 GPUs, two AMD EPYC 7543 32-Core CPUs, and 504GB of memory. The linkage of all the experiments is performed on another machine with two Intel Xeon Gold 6248R CPUs and 377GB memory. Additional 500GB swap space is needed to link large datasets like \texttt{song}.

\paragraph{Licenses.} Our codes are based on Python 3.8 and some public python packages. No existing codes are included in FedSim. We will release our codes under Apache V2 license\footnote{\url{https://www.apache.org/licenses/LICENSE-2.0}}.

The datasets that we use in our experiments have different licenses as summarized in Table~\ref{tab:data_license}. All of them can be used for analysis but only some of them can be shared or used commercially.
\begin{table}[htpb]
\caption{Licenses of datasets} \label{tab:data_license}
\begin{threeparttable}
    \centering
    \begin{tabular}{cc cccc}
    \toprule
        \textbf{License} & \textbf{Dataset} & \textbf{Analyze} & \textbf{Adapt} & \textbf{Share} & \textbf{Commercial} \\
        \midrule
        CC BY 4.0\tnote{a} & \cite{steam,company} & \cmark & \cmark & \cmark & \cmark  \\
        CC BY-NC-SA 4.0\tnote{b} & \cite{house} & \cmark & \cmark & \cmark & \xmark  \\
        CC BY-SA 4.0\tnote{g} & \cite{loan} & \cmark & \cmark & \cmark & \cmark \\
        CC0 1.0\tnote{c} & \cite{airbnb,school,boone,frog}& \cmark & \cmark & \cmark & \cmark  \\
        Singapore Open Data License\tnote{d} &  \cite{hdb} & \cmark & \cmark & \cmark &\xmark \\
        NYCBS Data Use Policy\tnote{e} & \cite{bike}& \cmark & \cmark & \cmark & \cmark \\
        All rights reserved & \cite{rawg,taxi}& \cmark & \xmark & \xmark & \xmark \\ 
        \bottomrule
    \end{tabular}
    \begin{tablenotes}
    \item[a] \url{https://creativecommons.org/licenses/by/4.0/}
    \item[b] \url{https://creativecommons.org/licenses/by-nc-sa/4.0/}
    \item[c] \url{https://creativecommons.org/publicdomain/zero/1.0/}
    \item[d] \url{https://data.gov.sg/open-data-licence}
    \item[e] \url{https://www.citibikenyc.com/data-sharing-policy}
    \item[f] \url{http://millionsongdataset.com/faq/}
    \item[g] \url{https://creativecommons.org/licenses/by-sa/4.0/}
    \end{tablenotes}
\end{threeparttable}
\end{table}
 
\section{Additional Experiments}\label{apdx:exp}
\subsection{Additional Experiment on Privacy}\label{subsec:exp_company}
\begin{figure}[htpb]
    \centering
    \includegraphics[width=0.4\linewidth]{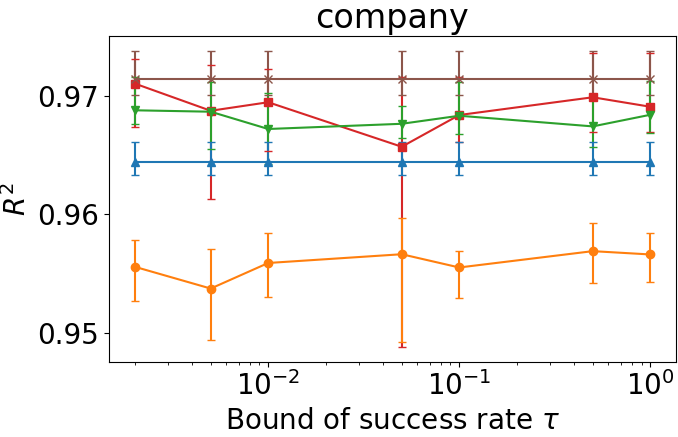}
    \caption{Performance with different scale of noise on similarities}
    \label{fig:perturb_company}
\end{figure}

Due to the page limit, we present the experiment of Section~\ref{sec:privacy} on \texttt{company} in Figure~\ref{fig:perturb_company}. We can make similar observations to Section~\ref{sec:privacy} that FedSim is robust to noise on similarities. 

\subsection{Effect of Hyperparameter K}\label{apdx:choose_k}

\begin{figure*}[htpb]
    \centering
    \includegraphics[width=.32\textwidth]{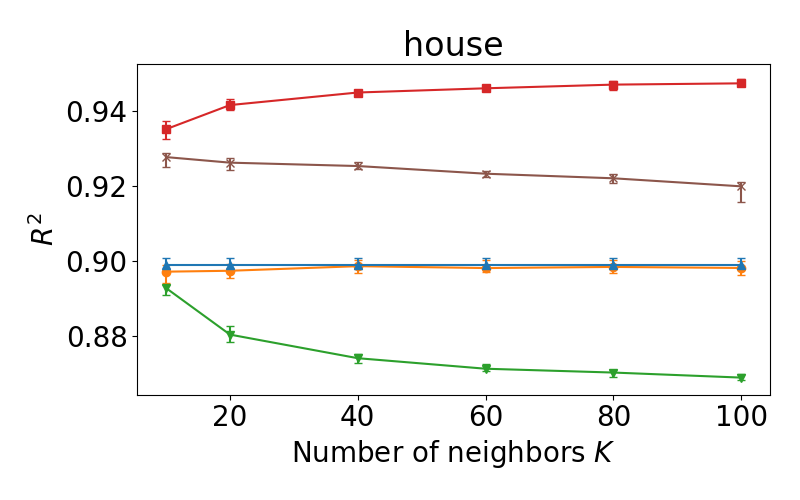}
    \includegraphics[width=.32\textwidth]{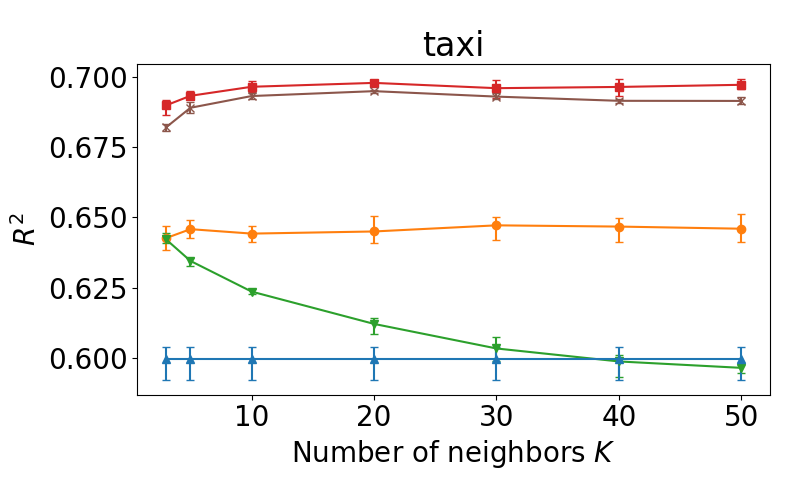}
    \includegraphics[width=.32\textwidth]{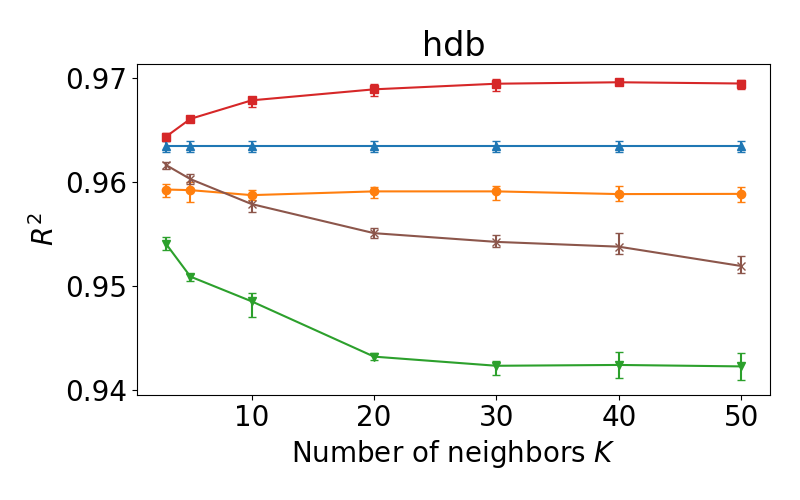}
    \includegraphics[width=.32\textwidth]{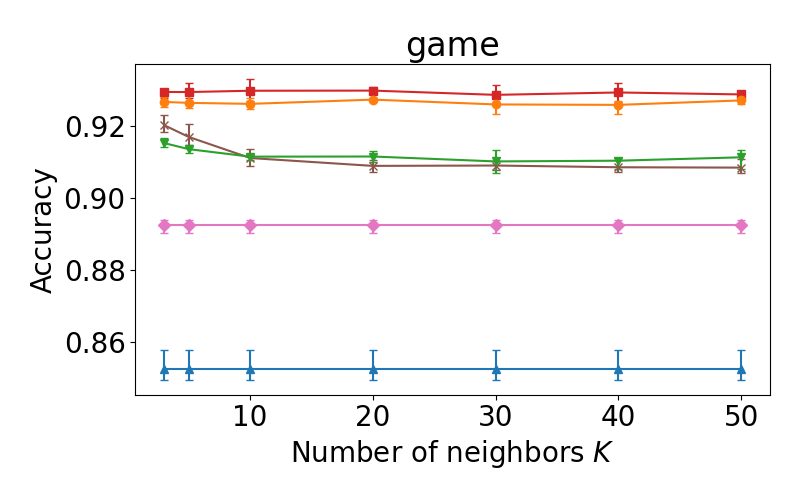}
    \includegraphics[width=.32\textwidth]{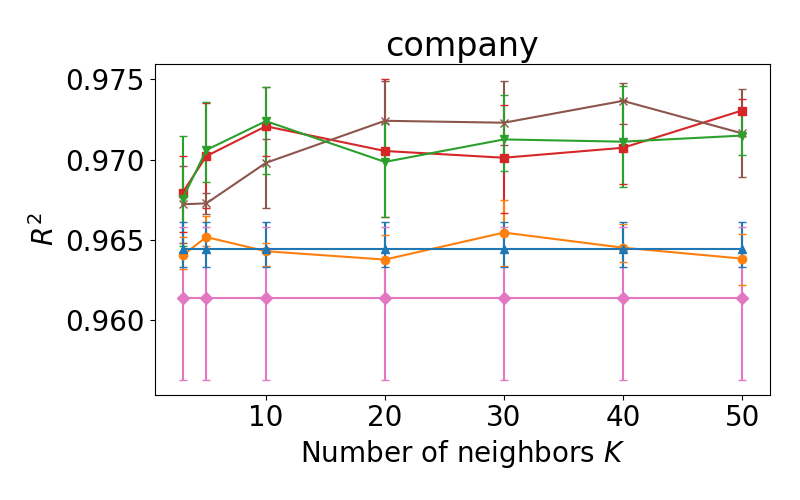}
    \includegraphics[width=0.5\linewidth]{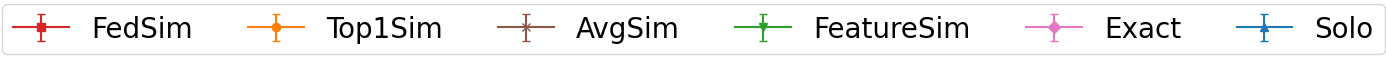}
    \caption{Performance with different $K$}
    \label{fig:exp_knn}
\end{figure*}

In this subsection, we aim to illustrate that the baselines cannot achieve good performance even by carefully tuning the number of neighbors $K$, while FedSim remains stable with large $K$ and consistently outperforms the baselines. With other hyperparameters fixed, we present the performance under different $K$ in Figure~\ref{fig:exp_knn}.

As we expect, the performance of Top1Sim remains steady since the pairs with the largest similarities remain the same. The performance of AvgSim and FeatureSim drops as $K$ increases since they cannot handle the redundant data properly. On the contrary, with increasing $K$, the performance of FedSim increases and then remains steady, which means FedSim can effectively exploit useful information in these additional data. In conclusion, \textbf{FedSim is robust to large $K$}; setting a relatively large $K$ achieves the best performance for FedSim.

\subsection{Multi-Party FedSim}\label{subsec:exp_multi}

In this subsection, we evaluate FedSim in multi-party setting on \texttt{house}. Specifically, we equally split the dataset for secondary parties into three parties by features. In Table~\ref{tab:multi-fedsim}, we report the performance when the party P conducts VFL with each combination of three secondary parties $\{S_1,S_2,S_3\}$. It can be observed that FedSim with more secondary parties produces better performance, which shows that FedSim can effectively utilize the data of more parties.

\begin{table}[htpb]
    \centering
    \small
    \caption{Performance of Four-Party FedSim on house}
    \label{tab:multi-fedsim}
    \begin{tabular}{ccc cc}
    \toprule
        \textbf{Secondary parties} & $S_1, S_2, S_3$ & $S_1, S_2$ & $S_1, S_3$ & $S_2, S_3$  \\ \midrule
        $R^2$ & 0.9475 & 0.9470 & 0.9481 & 0.9456  \\ \midrule
         \textbf{Secondary parties} & $S_1$ & $S_2$ & $S_3$ & Solo \\ \midrule
        $R^2$ & 0.9469 & 0.9114 & 0.9457 & 0.8989 \\
        \bottomrule
    \end{tabular}
    \vspace{-6pt}
\end{table}

\subsection{Time Consumption of FedSim} \label{subsec:train_time}
\paragraph{Linkage.} \wzmb{We present the time for linkage in Table~\ref{tab:linkage_time}. All the datasets organically contain either float or string identifiers. For float identifiers, Euclidean similarities are calculated; for string identifiers, edit similarities are calculated. Meanwhile, all the identifiers can be converted to Bloom filters according to \cite{karapiperis2017federal} for privacy. The Hamming similarities between Bloom filters are calculated. As can be observed from Table~\ref{tab:linkage_time}, linking Bloom filters for privacy is generally more time-consuming than linking raw float/string features. This indicates that a more efficient PPRL approach is desired. We leave this topic as our future study.}

\begin{table}[htpb]
    \centering
    \caption{\wzmb{Time for record linkage (min)}}
    \label{tab:linkage_time}
    \begin{tabular}{cccccc}
        \toprule
        \textbf{Type of Identifiers} & \textbf{house} & \textbf{taxi} & \textbf{hdb} & \textbf{game} & \textbf{company} \\ \midrule
        Float/String & 3.48 & 12.95 & 0.98 & 3.62 & 13.20 \\
        Bloom Filter \cite{karapiperis2017federal} & 145.20 & 594.27 & 258.23 & 41.38 & 19.87 \\
        \bottomrule
    \end{tabular}
    
\end{table}

\paragraph{Training.} FedSim has more parameters compared with other one-to-one baselines, thus leading to a longer training time. For each dataset, we report the training time per epoch in Table~\ref{tab:train_time}\wzmb{, and present the number of parameters of each model in Table~\ref{tab:n_params}}. \wzm{Note that the soft linkage procedure (including similarity calculation and k-nearest-neighbor search), as a preprocessing step, is not included in the training time.}

Two observations can be made from Table~\ref{tab:train_time}. \wzm{First, FedSim, AvgSim, and FeatureSim require \textbf{longer but acceptable training time} compared to other baselines. This is because these three approaches train $K$ times more samples than other approaches, thus costing approximately $K$ times longer time. Second, FedSim requires \textbf{similar training time} compared to AvgSim and FeatureSim in each epoch.} \wzmb{Although FedSim contains around $K$ times more parameters (mostly in merge model) than AvgSim and FeatureSim, the trained samples in the merge model are also $K$ times fewer than the original batch size, because each $d_i^P$, together with its $K$ neighbors $\mathbf{d}_i^S$, is regarded as one 2D sample in the merge model. Therefore, the per-epoch training time of FedSim is similar to that of AvgSim and FeatureSim.}

\begin{table}[htpb]
    \centering
    \caption{Training time (seconds) per epoch of different VFL algorithms on real-world datasets}
    \label{tab:train_time}
    \begin{tabular}{cccccc}
        \toprule
        \textbf{Models} & \textbf{house} & \textbf{taxi} & \textbf{hdb} & \textbf{game} & \textbf{company} \\
        \midrule
        Exact & - & - & - & $<$1 & 4 \\
        FeatureSim & 15 & 51 & 5 & 1 & 23 \\
        AvgSim & 6 & 35 & 4 & 1 & 15 \\
        Top1Sim &  $<$1 & 1 & $<$1 & $<$1 & 3 \\
        FedSim & 9 & 38 & 6 & 4 & 62 \\
        \bottomrule
    \end{tabular}
\end{table}

\begin{table}[htpb]
    \centering
    \caption{\wzmb{Number of parameters of each model}}
    \label{tab:n_params}
    \begin{tabular}{cccccc}
    \toprule \textbf{Models} & \textbf{house} & \textbf{taxi} & \textbf{hdb} & \textbf{game} & \textbf{company} \\
    \midrule
Exact&-&-&-&63,301&85,401\\
Featuresim&76,201&116,801&76,201&32,801&45,201\\
Avgsim&76,001&116,701&136,601&63,301&75,701\\
Top1sim&76,001&11,971&76,001&63,301&170,601\\
Fedsim&3,469,806&1,846,490&1,869,806&1,826,506&494,474\\

\bottomrule
    \end{tabular}
\end{table}

\subsection{Visualization of the Similarity Model and Merge Model}\label{subsec:visualization}
\begin{figure}[htpb]
    \centering
    \includegraphics[width=.9\linewidth]{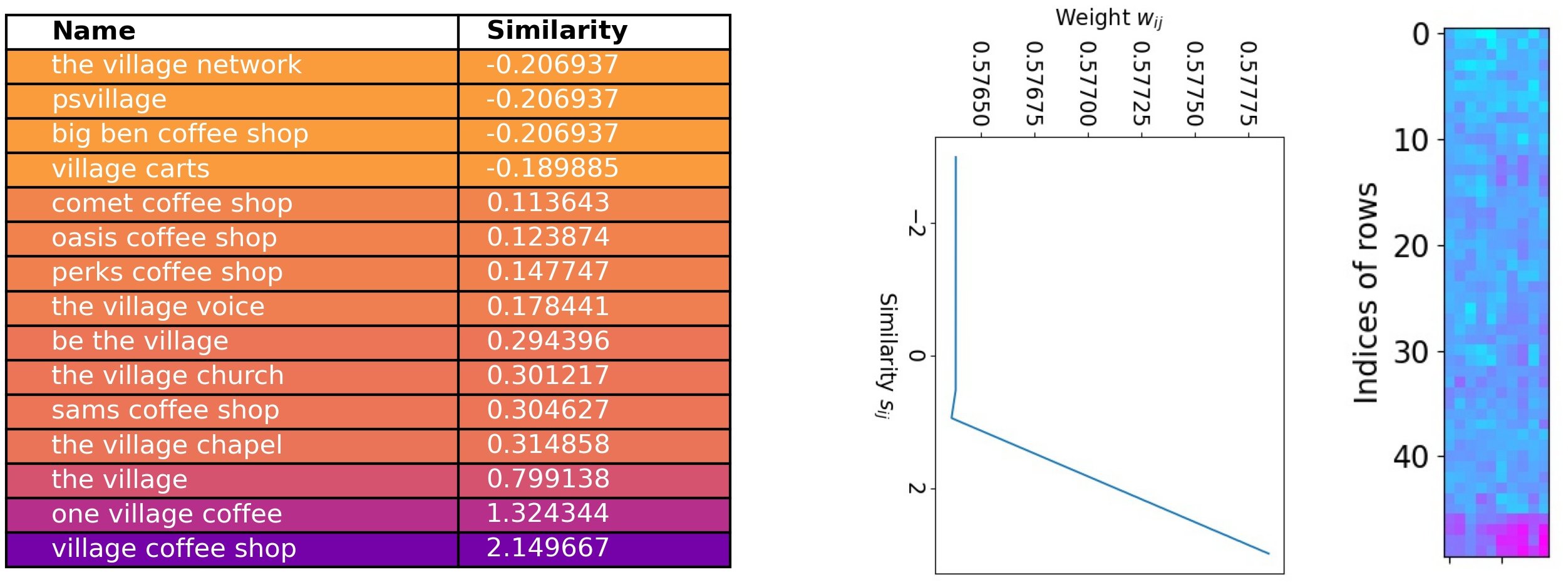}
    \caption{\wzmc{Visualization of FedSim on a data batch of \texttt{company} dataset. The left figure includes identifiers (``company name'') similar to ``the village coffee shop''. The middle figure shows the mapping from similarities to weights in the similarity gate. The right figure displays the feature importance calculated by \textit{integrated gradients} \cite{sundararajan2017axiomatic} of each input feature $(50\times10)$ of merge gate.}}
    \label{fig:insight}
\end{figure}

\wzmd{To provide more insights into FedSim, we visualize the similarity model and the merge model as Figure~\ref{fig:insight}, both of which are extracted from the converged FedSim on \texttt{company} dataset. For the data records linked with ``the village coffee shop'' (left), we highlight similar records with deeper colors. For the similarity model (middle), which has one-dimensional input and one-dimensional output, we plot the model in a two-dimensional coordinate system when scaled similarities range in [-3,3]. For the merge model (right), we evaluate the feature importance of each feature in $\mathbf{o}_t'(50\times10)$ by \textit{integrated gradients} \cite{sundararajan2017axiomatic}, which is widely used \cite{qi2019visualizing,sayres2019using} in model explanation. Larger vertical indices indicate larger similarities due to the sort gate. The deeper colors indicate higher importance and the lighter colors indicate lower importance.}

\wzmd{Three observation can be made from Figure~\ref{fig:insight}. First, ``the village coffee shop'' is soft-linked with three categories of companies: (a) exactly matched ``village coffee shop'', (b) other coffee shops, and (c) other companies unrelated to the coffee shop (e.g., the village network). Intuitively, (a) and (b) tends to have more similar feature values (e.g., number of employees) with ``the village coffee shop'' than (c). Therefore, when training ``the village coffee shop'', the linked pairs with (a) and (b) should have a larger effect than (c). Second, as expected, in the similarity model, the record in (a) is granted a dominant weight, the records in (b) are granted moderate-to-high weights, and the records in (c) are granted small weights. Third, in the merge model, the linked records with high similarities, as an input feature for the merge model, have a more significant effect on the merge model. These observations imply that the weight gate and merge gate can effectively capture the key data records and filter useless records according to the similarities.}

\subsection{Independent Common Features}\label{apdx:random_feature}
In this subsection, we study a special case where there is no mutual information between common features and unique features. We generate a synthetic dataset \texttt{sklearn-random} similarly to \texttt{sklearn} dataset. The only difference is that \texttt{sklearn-random} generates random common features, while \texttt{sklearn} selects existing features as common features. In this case, the similarity between common features is independent of the other features. Thus, we expect FedSim to perform the same as Top1Sim/Exact baseline at a reasonable $K$. After training FedSim with CNN merge model five times, the results are summarized in Table~\ref{tab:random_feature}.

\begin{table}[htbp]
    \centering
    \small
        \caption{Performance on random common features}
    \label{tab:random_feature}
    \begin{tabular}{c c}
    \toprule
        \textbf{Method} & \textbf{Accuracy} \\
        \midrule
        Solo & 60.76$\pm$0.41\\ 
        Top1Sim/Exact & 92.15$\pm$0.33 \\
        Combine &94.70$\pm$ 0.58 \\
        FedSim ($K=50$)  & 87.79$\pm$ 0.77 \\
        \textbf{FedSim ($K=30$)} & \textbf{92.80$\pm$ 0.61} \\
        \bottomrule
    \end{tabular}

\end{table}

From Table~\ref{tab:random_feature}, we observe that the performance of FedSim is similar to Exact and Combine when $K = 30$ though dropping lower when $K = 50$ due to the noise. This demonstrates the basic assumption behind FedSim: \textbf{the similarity between identifiers is related to the similarity between records}. FedSim outperforms baselines when the relationship is tight (e.g. identifiers are GPS), and provides similar performance to baselines when the relationship is weak or non-existent (e.g. identifiers are hash value). Notably, by properly setting hyperparameters, FedSim will not degrade under baselines because \texttt{Exact} and Top1Sim are special cases of FedSim when setting the number of neighbors $K$ to 1 with a linear merge model.

\subsection{Different Similarity Metrics}
\wzmb{In this subsection, we study how different similarity metrics affect the performance of FedSim. For numeric identifiers, we test Euclidean similarity and Hamming similarity; for string identifiers, we test edit similarity and Hamming similarity. The performance of FedSim is presented in Table~\ref{tab:sim_metric}. We observe that the dataset with Hamming similarity is usually lower than other metrics because generating Bloom filters introduces random noise. The only exception is the \texttt{company} dataset with long strings of company names as identifiers. In these names, words can be dislocated or shuffled, thus failing the edit similarity. For example, ``Kentucky bank'' and ``bank of Kentucky'' with a large edit distance can be recognized as similar by Bloom filters built from q-grams. In general, the similarity metric does not significantly affect the performance.}

\begin{table}[htpb]
    \centering
    \small
    \setlength\tabcolsep{3pt}
    \caption{\wzmb{Performance of FedSim on different similarity metrics}}
    \label{tab:sim_metric}
    \begin{tabular}{cccc ccc}
    \toprule
    \multirow{1}{*}{\textbf{Similarity Metric}} & \multicolumn{1}{c}{\textbf{house (numeric)}} & \multicolumn{1}{c}{\textbf{bike (numeric)}} & \multicolumn{1}{c}{\textbf{hdb (numeric)}} & \textbf{game (string)} & \multicolumn{1}{c}{\textbf{company (string)}} \\
    \midrule
    Euclidean/Edit & 42.12\textpm 0.23 & 235.67\textpm 0.27 & 27.13\textpm 0.06 & 92.88\textpm 0.11\%&40.04\textpm 2.18 \\
    Hamming & 50.03\textpm 1.09&238.81\textpm 0.50&28.29\textpm 0.22&92.70\textpm 0.48\%&37.08\textpm 0.61\\
        \bottomrule
    \end{tabular}
    
\end{table} 

\section{Limitations}\label{sec:limitation}
\wzm{We discuss the limitation of FedSim in the following two aspects.}

\paragraph{Accuracy.} \wzmc{As discussed in Appendix~\ref{apdx:application} and Appendix~\ref{apdx:random_feature}, the performance boosting of FedSim is based on an assumption that \textit{the similarity between identifiers is related to the similarity between data records}. This assumption may not hold in a small portion of real applications (e.g., identifiers are randomly generated IDs). In these cases, FedSim has no improvement on baselines and can even be outperformed by baselines if $K$ is too large (see Appendix~\ref{apdx:random_feature}). Further improvements on these cases are left as our future work.}

\paragraph{Privacy.} \wzm{As stated in Section~\ref{sec:privacy}, we propose an intuitive attack and the corresponding defense. Nonetheless, this approach does not fully guarantee the privacy of FedSim. We further discuss three directions of advanced attacks in Appendix~\ref{sec:adv_attack}. The design of these advanced attacks is non-trivial, thus we leave these advanced attacks as well as their defense as our future work.}

\paragraph{Scalability.} \wzm{In Appendix~\ref{sec:multi_party}, we propose an extension of FedSim to the multi-party setting under the assumption that the primary party holds more important or comparable features than secondary parties. In this case, FedSim is scalable on parties since secondary parties can be linked by one-to-one mappings without significant performance loss. Nevertheless, if this assumption no longer holds, performing one-to-many linkage between each pair of parties is infeasible due to the expensive computational cost. Designing a non-exact VFL model for multiple parties remains an open problem in VFL.}

\section{Discussion}\label{sec:discussion}
\paragraph{Potential negative societal impact.} \texttt{FedSim} might be adapted to a linkage attack method, which could threaten the privacy of released data. Since we provide an effective approach to exploit information through fuzzy matched identifiers (e.g. GPS locations), this kind of identifier should be paid special attention to privacy when releasing a dataset.

\paragraph{Consent of dataset.} As discussed in Appendix~\ref{sec:exp_detail}, all the datasets that we are using are publicly available for analysis. Therefore, we naturally obtain consent of using these datasets by following their licenses.

\paragraph{Personally identifiable information.} To the best of our knowledge, the datasets that we are using do not contain personally identifiable information or offensive contents.

\paragraph{Extension to other VFL algorithms} FedSim makes a significant step toward practical VFL by enabling VFL on a wider range of applications that requires fuzzy matching. Although \texttt{FedSim} is designed based on SplitNN, the idea of working directly with similarities can also be developed in other VFL frameworks. \wzmb{Unfortunately, the framework of FedSim cannot be directly applied to other VFL algorithms, e.g., logistic regression and tree-based algorithms. This is because FedSim requires the similarity model and merge model to be trained together with the main VFL model, which requires the VFL algorithm to be neural-network-based. To adapt FedSim to other VFL algorithms, a possible direction is to exchange the intermediate information in these algorithms like exchanging gradients in SplitNN. We leave this topic as our future work.}

\paragraph{Relation to other topics.} \wzmc{Besides VFL, our main insight that \textit{the similarity between identifiers is related to the similarity between data records} has also been recognized in many other topics. For example, k-nearest neighbors (k-NN) algorithm \cite{peterson2009k} predicts by averaging the most similar samples. Graph neural networks (GNN) \cite{hamilton2017inductive} adopt a loss function that encourages neighboring nodes to have similar representations. Semi-supervised learning \cite{zheng2022simmatch} can benefit from matching similar instances. FedSim is the first approach that exploits this insight in VFL.} 

\section{Additional Background}\label{apdx:background}
FEDERAL \cite{karapiperis2017federal} is a PPRL framework that theoretically guarantees the indistinguishability of bloom filters. Suppose the size of bloom filters is set to $N$. For strings, it generates q-grams and encodes q-grams to bloom filters of size $N$ by composite cryptographic hash functions. For numeric values, it first generates $N$ random numbers $r_i\,(i\in[1,N])$ and sets a threshold $t$. Next, it determines $N$ hash functions $h_i(x)$
\[
h_i(x)=
\begin{cases}
1, & x\in[r_i-t,r_i+t] \\
0, & otherwise
\end{cases}\,(i\in[1,N])
\]

Each numeric value is hashed by all the functions $h_i\,(i\in[1,N])$ and is converted to a bloom filter of size $N$. These bloom filters are used to calculate similarities on an honest-but-curious third party. They prove that all the bloom filters have similar numbers of ones if we properly set the size of bloom filters. Formally, denoting $\omega$ as the number of ones in a bloom filter, we have
\(
\Pr[\omega\le (1-\epsilon)E[\omega]]< \delta
\),
where $\epsilon<1$ and $\delta<1$ are tolerable deviations. Therefore, attackers cannot infer whether the numeric values are large or small given the bloom filters. This method is adopted in our experiment on hamming-based similarities.

\end{document}